\documentclass[journal]{IEEEtran}

\usepackage{times}
\usepackage{epsfig}
\usepackage{graphicx}
\usepackage{amsmath}
\usepackage{amssymb}
\usepackage{color}
\usepackage{subfigure}
\usepackage{algorithm}
\usepackage{algorithmic}
\usepackage{url}

\newcommand{\ignore}[1]{}
\usepackage{multirow}
\usepackage[table,xcdraw]{xcolor}
\begin{document}

\title{Vision-based Anti-UAV Detection and Tracking}

\author{Jie Zhao,
        Jingshu Zhang,
        Dongdong Li, 
        Dong Wang

}

\markboth{IEEE Transactions on Intelligent Transportation Systems}%
{Wang~\MakeLowercase{\textit{et al.}}: Vision-based Anti-UAV Detection and Tracking}

\maketitle

\begin{abstract}
Unmanned aerial vehicles (UAV) have been widely used in various fields, and their invasion of security and privacy has aroused social concern. Several detection and tracking systems for UAVs have been introduced in recent years, but most of them are based on radio frequency, radar, and other media. We assume that the field of computer vision is mature enough to detect and track invading UAVs. Thus we propose a visible light mode dataset called Dalian University of Technology Anti-UAV dataset, DUT Anti-UAV for short. It contains a detection dataset with a total of 10,000 images and a tracking dataset with 20 videos that include short-term and long-term sequences. All frames and images are manually annotated precisely. We use this dataset to train several existing detection algorithms and evaluate the algorithms' performance. Several tracking methods are also tested on our tracking dataset. Furthermore, we propose a clear and simple tracking algorithm combined with detection that inherits the detector’s high precision. Extensive experiments show that the tracking performance is improved considerably after fusing detection, thus providing a new attempt at UAV tracking using our dataset.
The datasets and results are publicly available at: \url{https://github.com/wangdongdut/DUT-Anti-UAV}.
\end{abstract}


\IEEEpeerreviewmaketitle

\section{Introduction}
\IEEEPARstart{W}{ith} the maturity of industrial technology, unmanned aerial vehicles (UAV) have gradually become mainstream. They are widely used in logistics~\cite{vskrinjar2018application}, transportation~\cite{xu2017car}, monitoring~\cite{cheng2017autonomous}, and other fields because of their small size, low price, and simple operation~\cite{lort2017initial}. Although UAVs provide convenience, they cause a series of problems. Either public safety or personal safety and privacy are easily violated. Therefore, the detection and tracking of illegally or unintentionally invading UAVs are crucial. However, no complete and reliable anti-UAV detection and tracking system is available at present. Most of the existing detection and early warning technologies are based on radar~\cite{hoffmann2016micro}, radio frequency (RF)~\cite{abunada2020design}, and acoustic sensors~\cite{chang2018surveillance}, which often have limitations, such as high cost and susceptibility to noise. These limitations lead to unreliable results. Therefore, these existing algorithms cannot be used extensively. Their application range is limited to airports and other public places.\par
In recent years, methods based on deep learning have developed rapidly in various fields~\cite{gao2020cross,gao2021constructing,gao2020hierarchical,ren2015faster,SiamFC} of computer vision, especially for object detection and tracking. Their maturity provides the possibility of establishing a high-performance tracking system of anti-UAV. Many generic object detection models, such as Faster-RCNN~\cite{ren2015faster} and SSD~\cite{liu2016ssd}, and common tracking models, such as SiamFC~\cite{SiamFC} and DiMP~\cite{dimp}, are currently available. However, these generic methods do not perform well when directly applied to UAV detection and tracking. Even though detection algorithms have gradually become mature and commercialized, small-target detection in the complex background is still a problem, which anti-UAV detection aims to address. UAV often fuses with the complex background with much noise and interference. Occlusion also occurs and brings challenges to the tracking task. A series of methods, such as improving YOLOv3~\cite{hu2019object}, which uses low-rank and sparse matrix decomposition to conduct classification~\cite{wang2019flying}, are proposed to solve the aforementioned problems, and achieve good results.\par
The main motivation of our work is to use existing state-of-the-art detection and tracking methods to effectively adapt and address the anti-UAV task in the data level and method level. First, deep learning-based methods require plenty of training data to obtain robust and accurate performance. Although several corresponding datasets are proposed, such as Anti-UAV~\cite{jiang2021anti} and MAV-VID~\cite{rodriguez2020adaptive}, they are still not enough to train a high-performance model. Therefore, to make full use of existing detection and tracking methods for the anti-UAV task in the data level, and promote further development of this area, we propose a visible light dataset for UAVs, including detection and tracking subsets. We also retrain several detection methods using our training set. Second, we attempt to further improve the UAV tracking performance at the method level. To be specific, we propose a fusion strategy to combine detection and tracking methods.\par
Our main contributions are summarized as follows.\par
\begin{itemize}
\item We propose an anti-UAV dataset called DUT Anti-UAV that contains detection and tracking subsets. The detection dataset includes a training set (5200 images), a validation set (2600 images), and a testing set (2200 images). The tracking dataset includes 20 sequences. It will be released publicly for academic research.
\item We evaluate state-of-the-art methods on our dataset, including 14 detectors and 8 trackers. Detectors are all retrained using the training set of our DUT Anti-UAV detection dataset.
\item A clear and simple fusion algorithm is proposed for the UAV tracking task. This algorithm integrates detection into tracking while taking the advantage of the detector’s high precision. Extensive experiments show that the tracking performance is improved significantly for most combinations of trackers and detectors.
\end{itemize}

\section{Related work}
\subsection{Object detection and tracking under the UAV view}
Different from Anti-UAV tasks, nowadays there are more discussions about object detection and tracking from the UAV view. Compared to cameras on moving vehicles, a UAV is more flexible for it is easy to control. Therefore, UAV is often used to realize aerial object tracking. Several UAV datasets have been constructed so far, e.g., UAV123~\cite{1uav123} for tracking, DroneSURF~\cite{2kalra2019dronesurf} and CARPK~\cite{3hsieh2017drone} for detection, and so on. Besides, several corresponding algorithms~\cite{4yu2020unmanned,6li2017visual,7xing2021siamese} have been proposed to address these two tasks. UAV detection and tracking are mostly overlooking from above, for which it gains large view scope. However, it brings new challenges, such as high density, small object and complex background. For these properties, Yu \emph{et al}.~\cite{4yu2020unmanned} consider contextual information using the Exchange Object Context Sampling(EOCS) method~\cite{5yu2018online} in tracking, to infer the relationships between the objects. To solve the problem of fast camera motion, Li \emph{et al}.~\cite{6li2017visual} optimize the camera motion model by projective transformation based on background feature points. Besides, Xing \emph{et al}.~\cite{7xing2021siamese} consider that in real time tracking, computing resources employed on UAVs are limited. To complement lightweight network, they propose a lightweight Transformer layer, and then integrate it into pyramid networks, thus finally build a real-time CPU-based tracker.\par
Aforementioned algorithms perform well on existing UAV tracking benchmarks, as well as promote the commercialization of aerial object tracking. UAV tracking is more and more popular and draw increasing attention, which makes the anti-UAV tracking essential as well.\par
\subsection{Anti-UAV methodology}
Safety issues derived from UAVs have elicited increasing in recent years. In particular, considering national security, many countries have invested much time and energy in researching and deploying quite mature anti-unmanned systems that are not based on deep learning in military bases. Universities and research institutions are continuously optimizing these anti-unmanned systems.\par
\textbf{ADS-ZJU}~\cite{shi2018anti}. This system combines multiple surveillance technologies to realize drone detection, localization, and defense. It deploys three sensors to collect acoustic signals, video images, and RF signals. The information is then sent to the central processing unit to extract features for detection and localization. ADS-ZJU uses a short-time Fourier transform to extract spectrum features of the received acoustic signals, and histograms of oriented gradients to describe the image feature. It also takes advantage of the characteristic that the spectrum of the UAV’s RF signal is different from that of the WiFi signal by using the distribution of RF signals’ strengths at different communication channels to describe the RF feature. After feature extraction, it utilizes support vector machine (SVM) to conduct audio detection, video detection, and RF detection in parallel. After that, the location of UAVs can be estimated via hybrid measurements, including DOA and RSS, under the constraints of the specific geographical area from video images. The use of multiple surveillance technologies complements the advantages and disadvantages of multiple technologies, so that the system has high accuracy. Meanwhile, it can conduct radio frequency interference which simple vision-based system cannot do. But in this system, each unit is scattered, which makes the system covers a large area, and its high cost also makes it not suitable for civil use.\par
\textbf{Dynamic coordinate tracing}~\cite{sheu2019development}. This study proposes a dual-axis rotary tracking mechanism, using a dual-axis tracing device, namely, two sets of step motors with a thermal imaging or full-color camera and sensing module to measure the UAV’s flight altitude. The device dynamically calculates the longitude and latitude coordinates in spherical coordinates. The thermal imaging and full-color cameras are optionally used under various weather conditions, making the system robust in different environments. This drone tracking device for anti-UAV systems is inexpensive and practical, however, its requirements for hardware facilities are still high.\par

\subsection{UAV dataset}
In addition to using other media to solve the problem of UAV detection, people have also begun to utilize deep learning based object tracking algorithms for UAV tracking due to the rapid development of computer vision in recent years. In the task of computer vision, dataset is an important factor in obtaining a model with strong robustness. Therefore, datasets for UAV detection and tracking have been proposed consistently. Several relatively complete existing UAV datasets are described below.\par
\textbf{MAV-VID}~\cite{rodriguez2020adaptive}. This is a dataset published by Kaggle in which UAV is the only detected object. It contains 64 videos (40,323 pictures in total), of which 53 are used for training and 11 are used for validation. In this dataset, the locations of the UAVs are relatively concentrated, and the differences between locations are mostly horizontal. The detected objects are small, the average size of which is 0.66\% of the entire picture. While in our dataset, the distribution of UAV is scattered, and the horizontal and vertical distribution are relatively more uniform, which makes the model trained by our dataset more robust.\par

\textbf{Drone-vs-Bird Detection Challenge}~\cite{coluccia2019drone}. This dataset is proposed in the 16-th IEEE International Conference on Advanced Video and Signal-based Surveillance (AVSS). As the name indicates, the prime characteristic of this dataset is that in addition to UAVs, a number of birds cannot be ignored in the pictures. The detector must successfully distinguish the drones and birds, alerting against UAV while not responding to the birds. However, the size, color, and even shape of the two may be similar, which brings challenges to the detection task. Different from the first version, this dataset adds land scenes in addition to sea scenes, which are shot by different cameras. \ignore{The UAVs are given full annotations but the birds are not, namely, a bird is not recognized as a class of the detected object.}Another characteristic of this dataset is that the size of the detected object is extremely small. \ignore{generally, the widths do not even exceed 3 pixels.} According to statistical analysis, the average size of the detected UAVs is $34\times23$ (0.1\% of the image size). Seventy-seven videos consist of nearly 10,000 images. In light of this situation, improving the algorithm with regard to this dataset, successfully reducing the high false positive rates, and further popularizing the method to other fields if it is robust make up the significance of this dataset. Scenes in such datasets are most seaside with a wide visual field. Different from them, we collect data mostly in the place with lots of buildings, which is more suitable for civilian use.\par
\begin{figure*}[htbp]
    \centering
    \subfigure[Detection-train]{\includegraphics[width=0.24\textwidth]{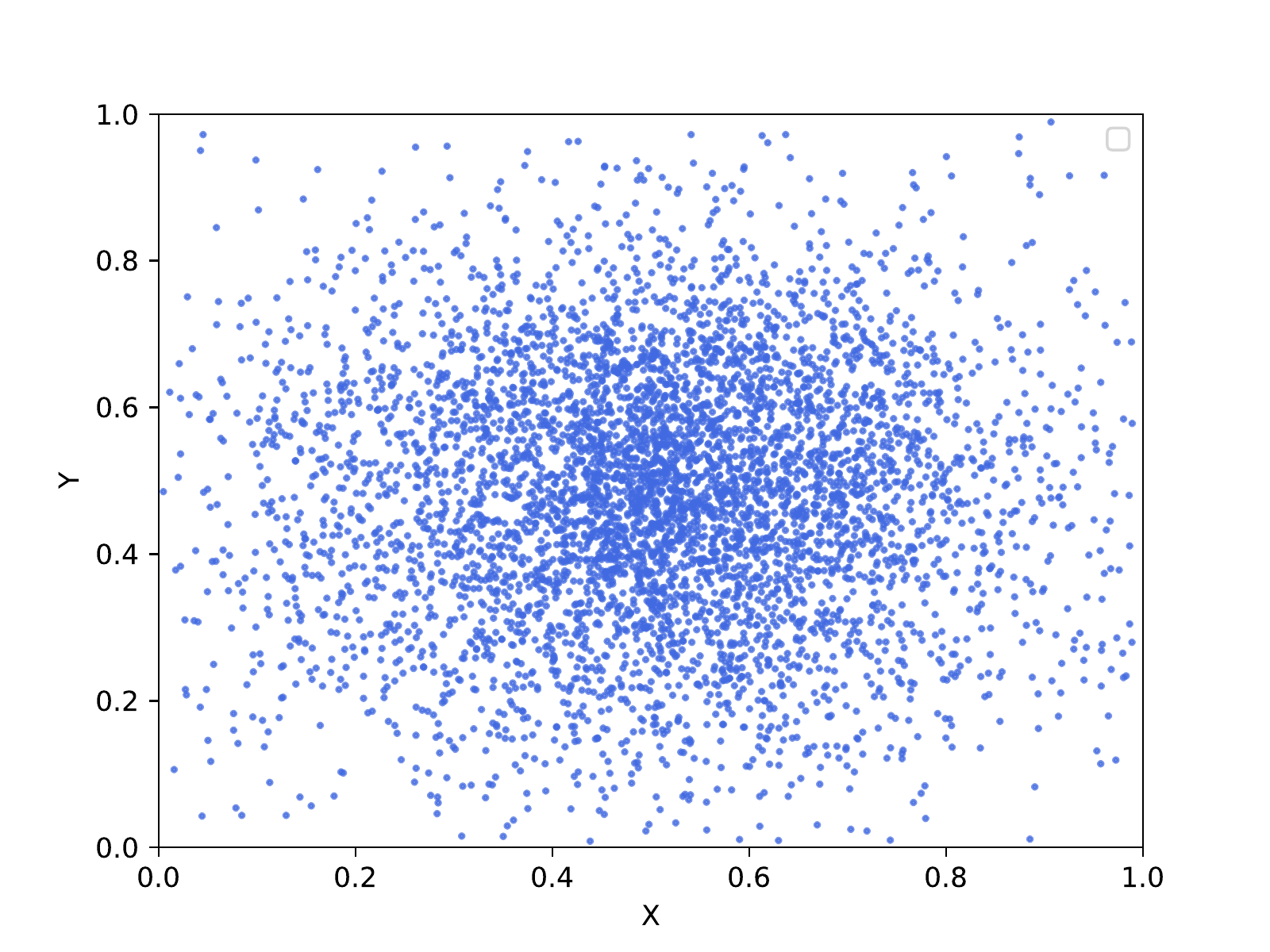}} 
    \subfigure[Detection-test]{\includegraphics[width=0.24\textwidth]{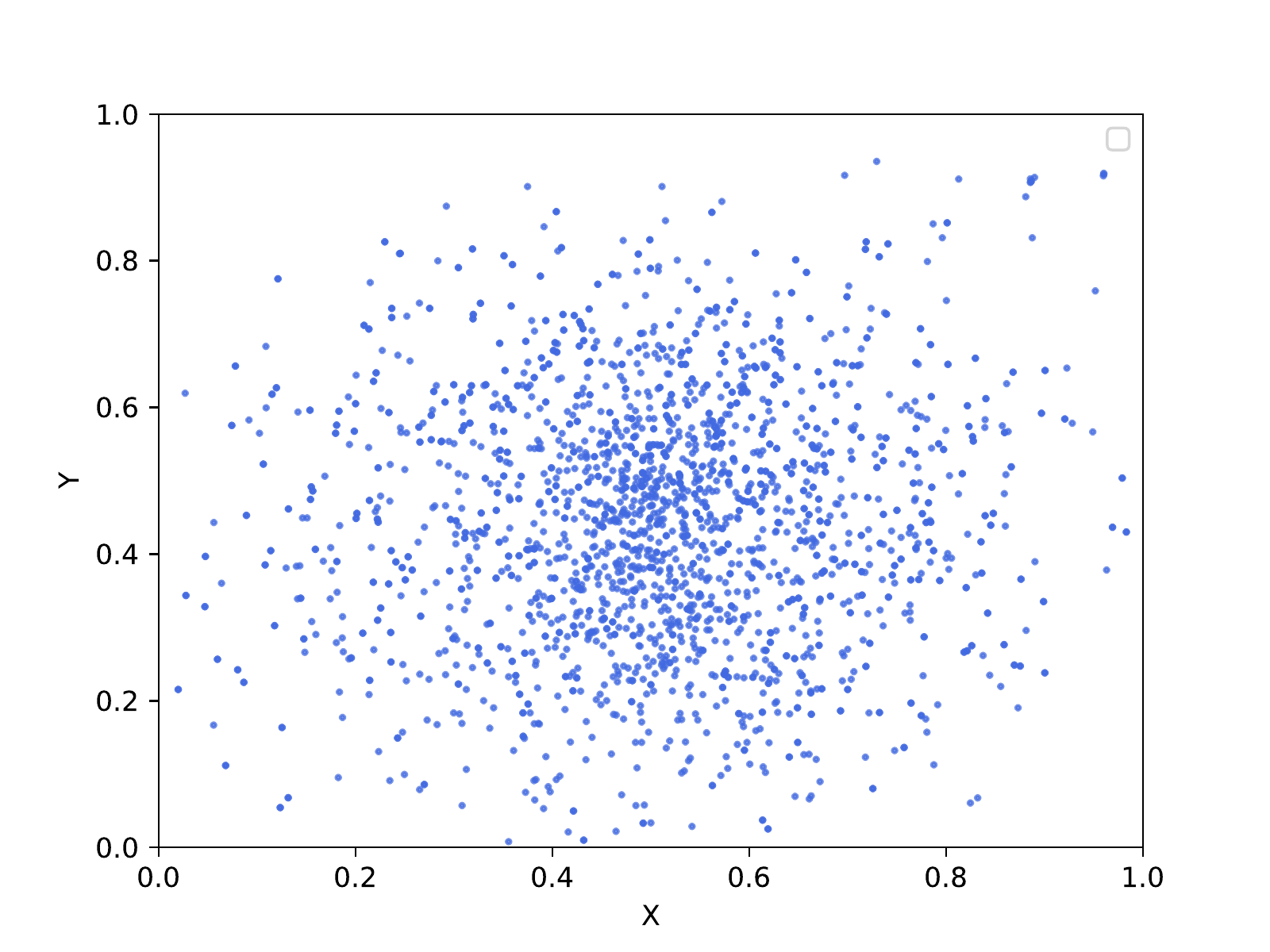}} 
    \subfigure[Detection-val]{\includegraphics[width=0.24\textwidth]{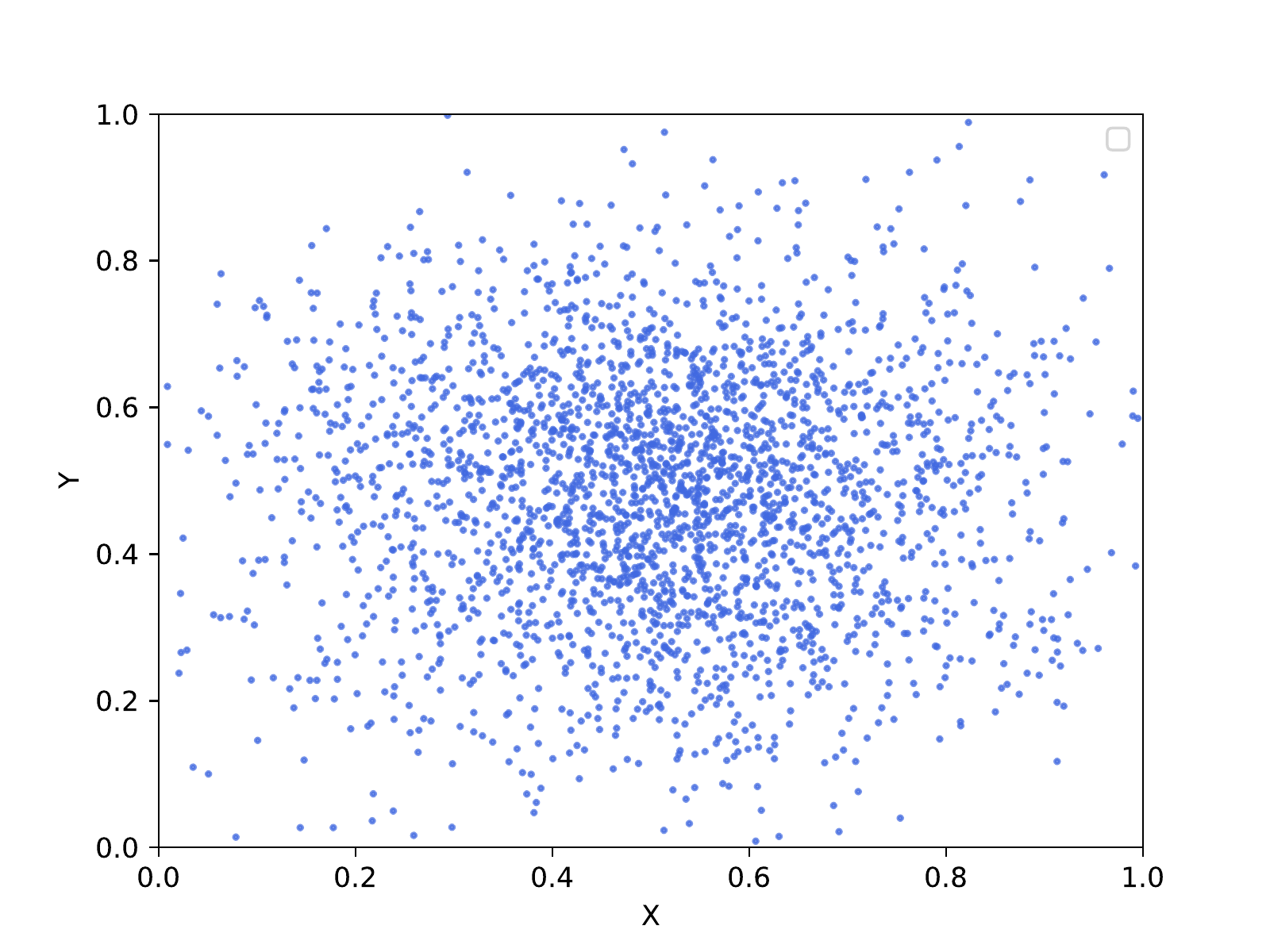}} 
    \subfigure[Tracking]{\includegraphics[width=0.24\textwidth]{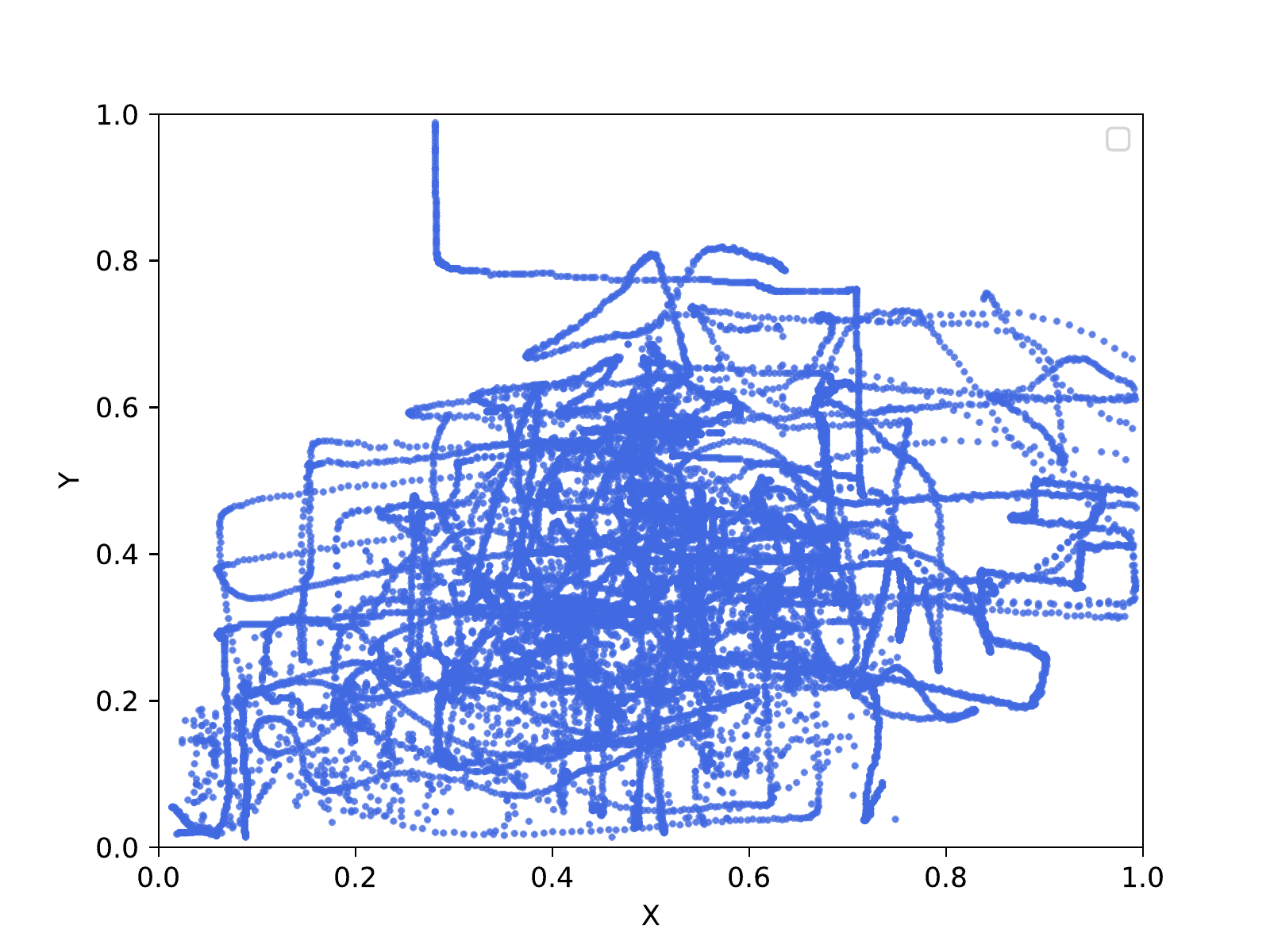}} 
    \caption{Position distribution of the DUT Anti-UAV dataset.}
    \label{fig:pos_dis}
\end{figure*}

\begin{figure*}[htbp]
    \centering
    \subfigure[Aspect ratio (Detection-train)]{\includegraphics[width=0.24\textwidth]{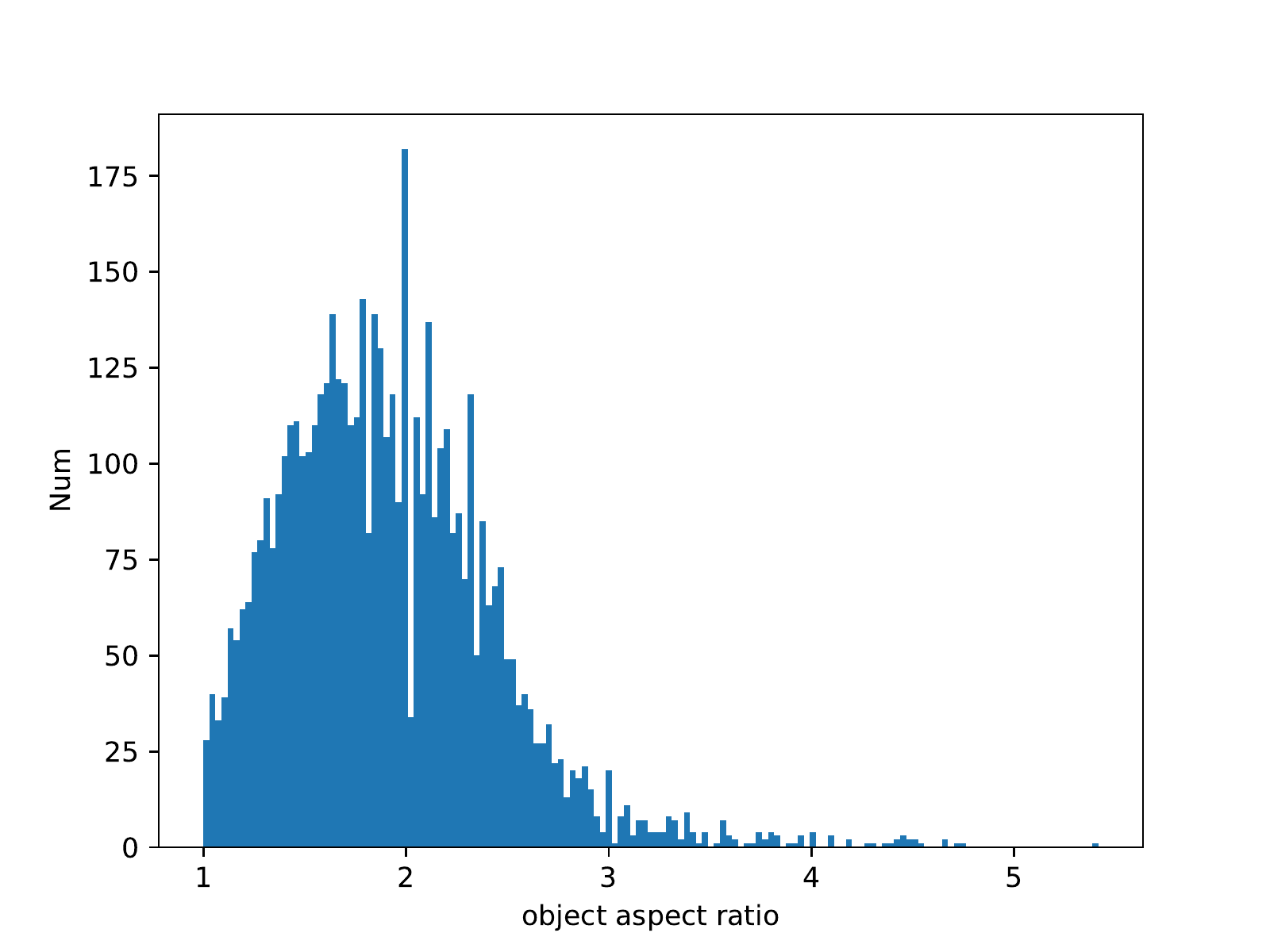}} 
    \subfigure[Aspect ratio (Detection-test)]{\includegraphics[width=0.24\textwidth]{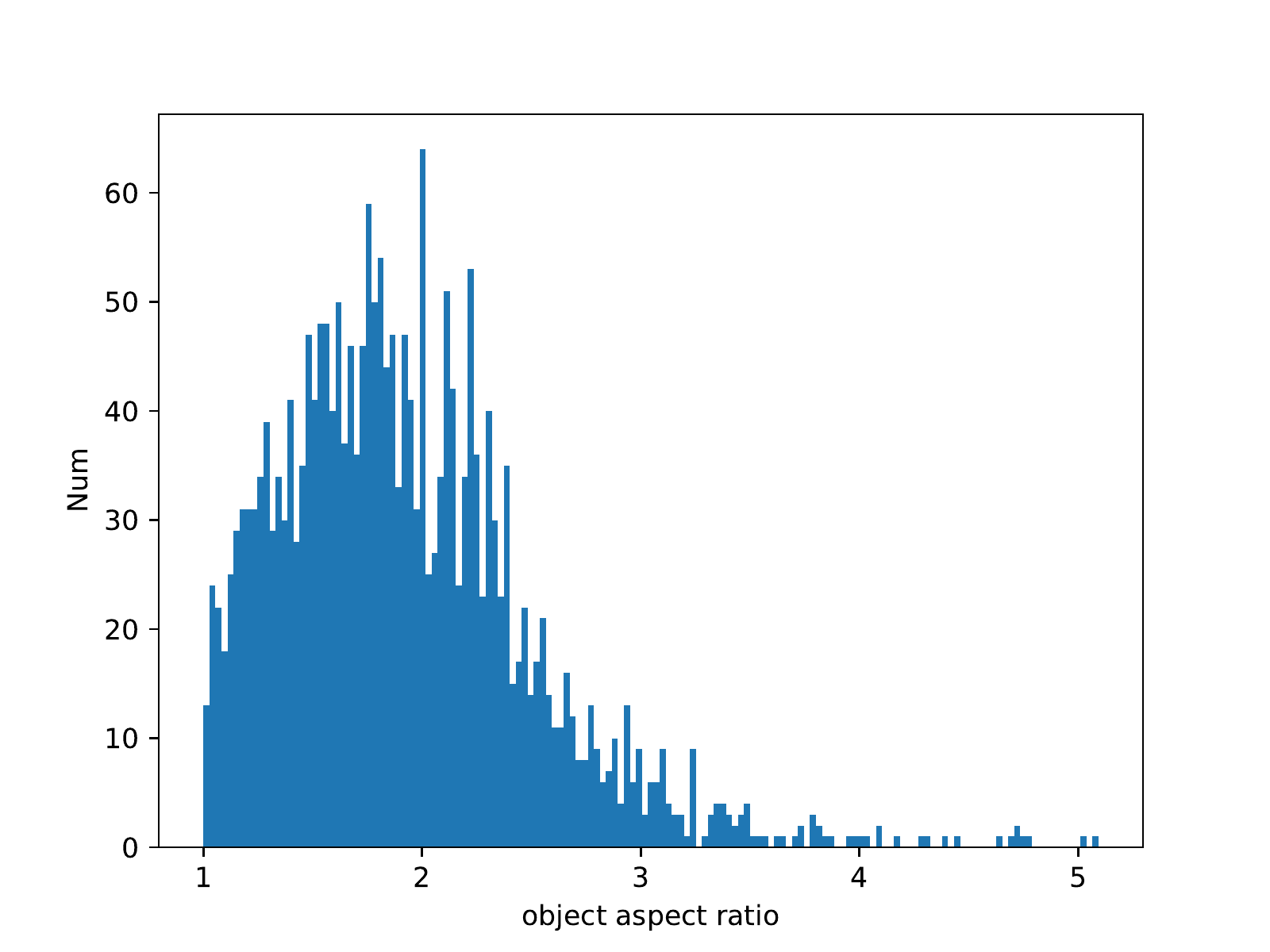}} 
    \subfigure[Aspect ratio (Detection-val)]{\includegraphics[width=0.24\textwidth]{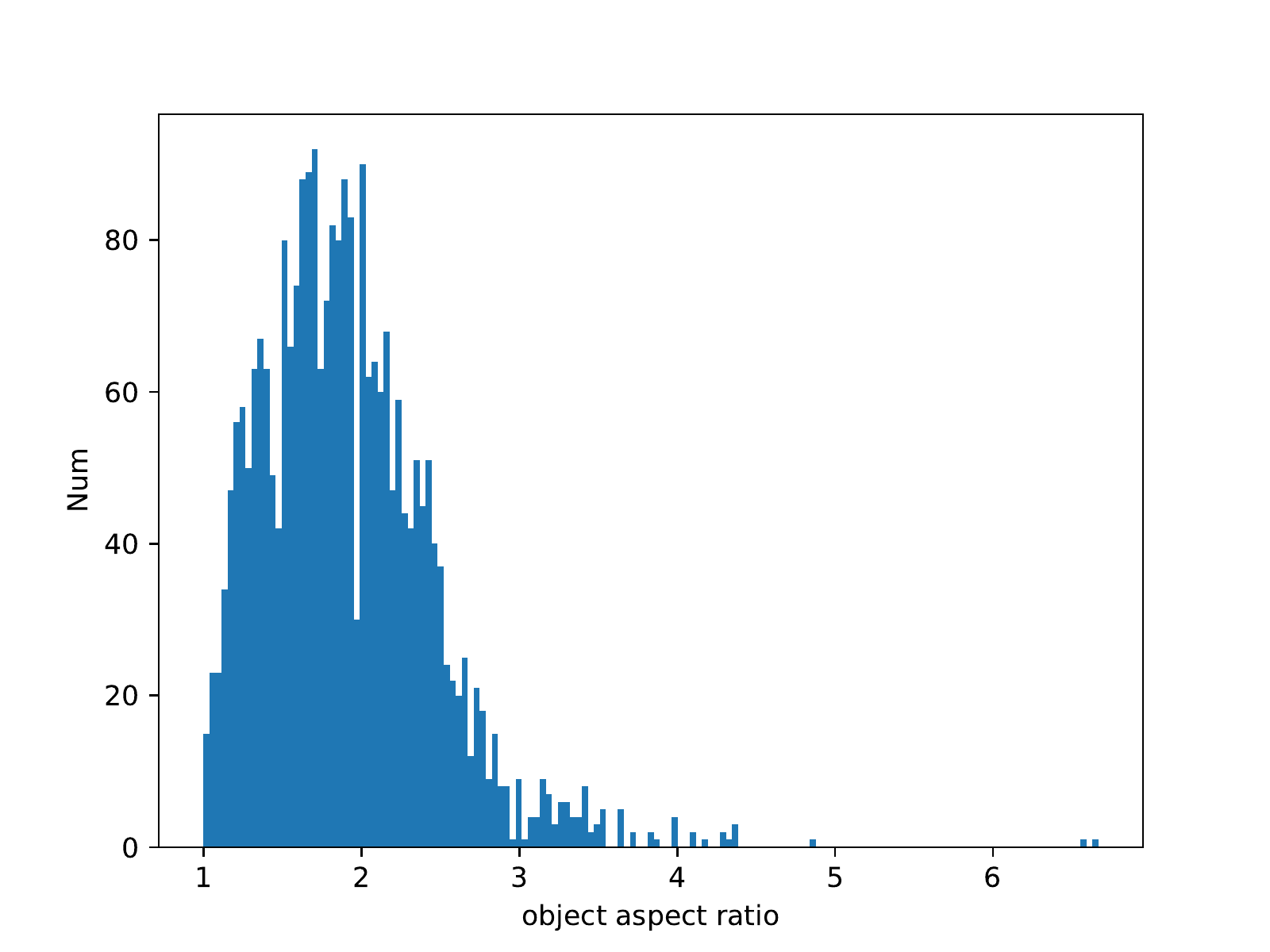}} 
    \subfigure[Aspect ratio (Tracking)]{\includegraphics[width=0.24\textwidth]{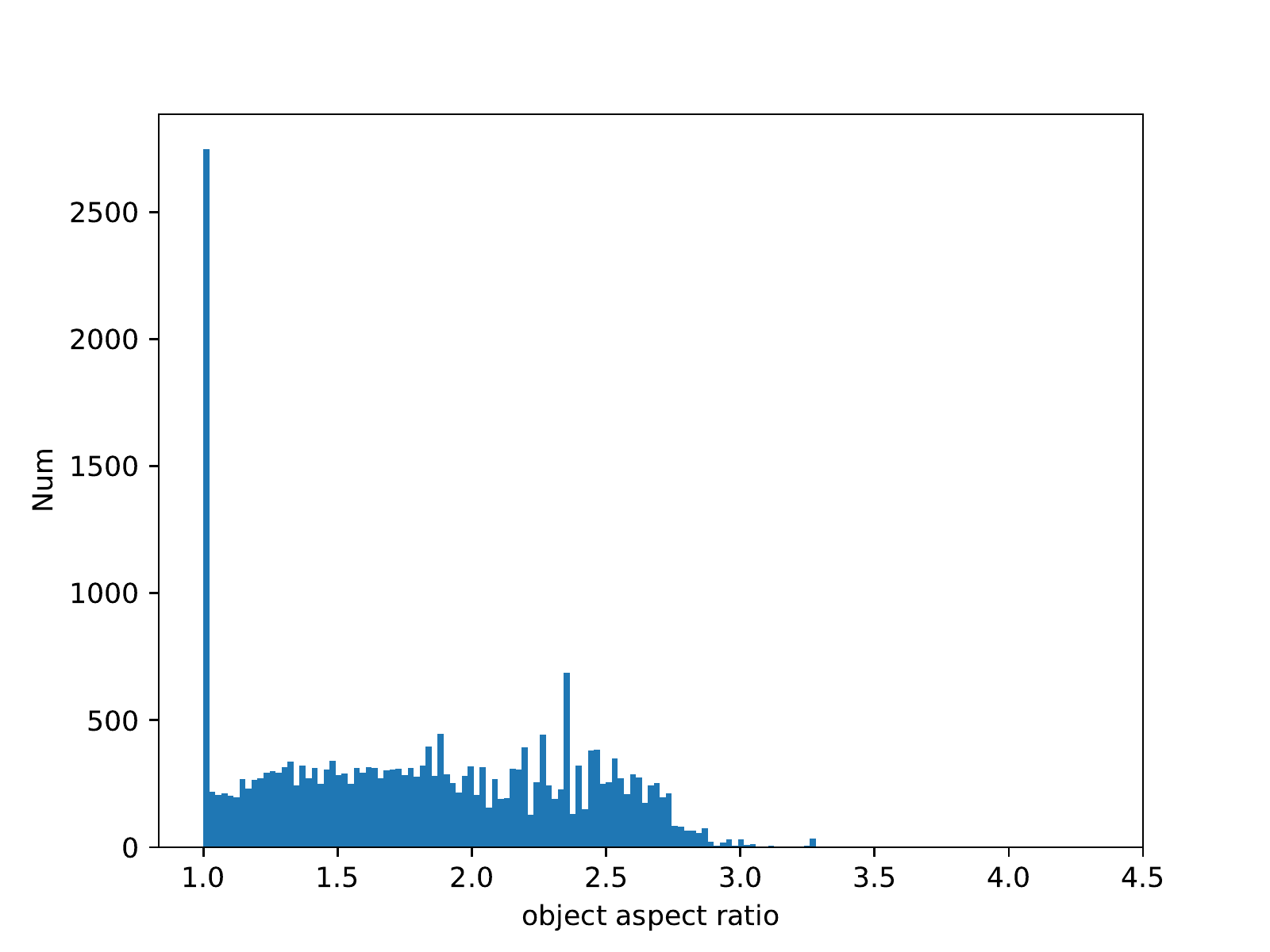}} \\
    \subfigure[Scale (Detection-train)]{\includegraphics[width=0.24\textwidth]{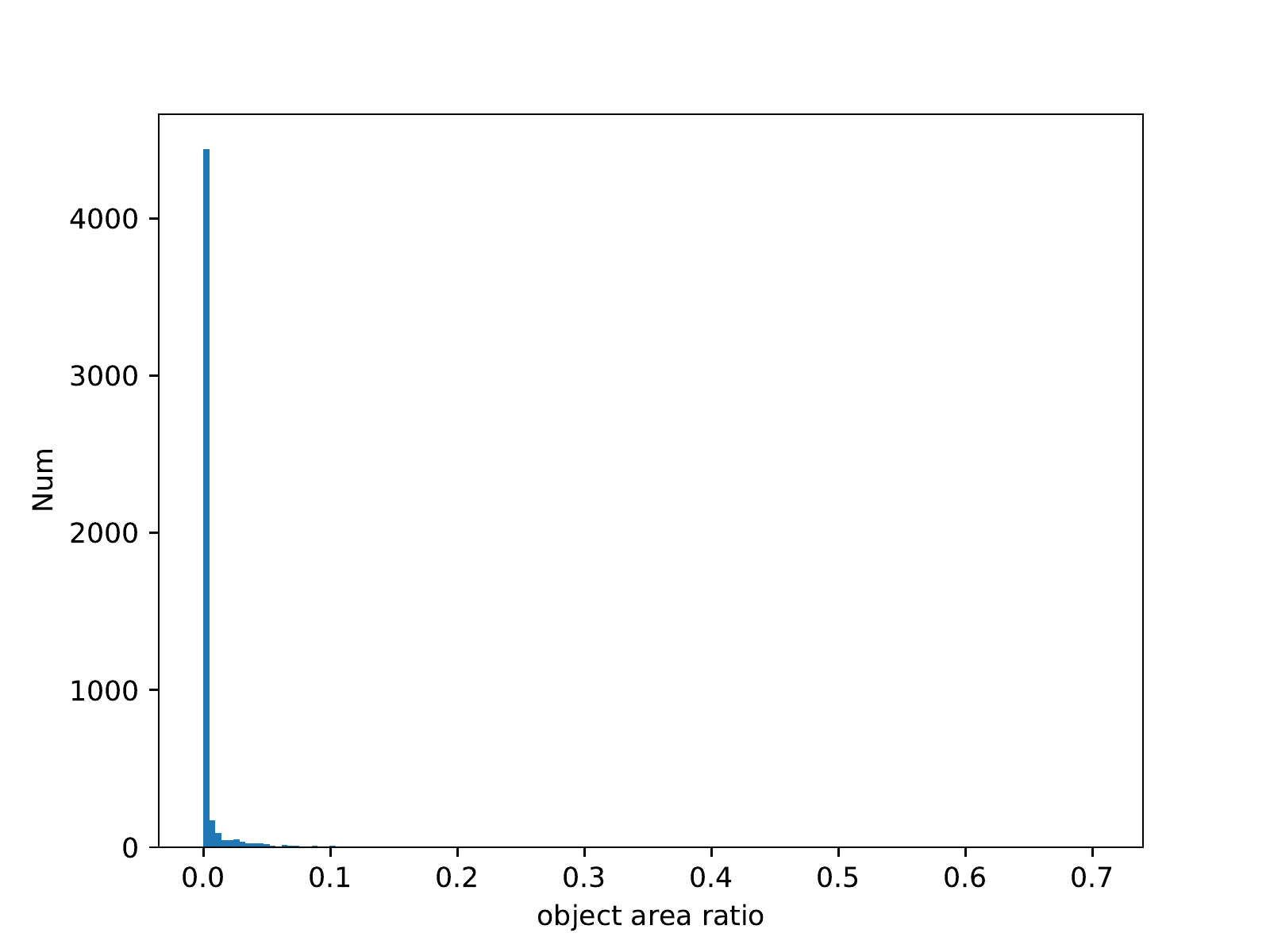}} 
    \subfigure[Scale (Detection-test)]{\includegraphics[width=0.24\textwidth]{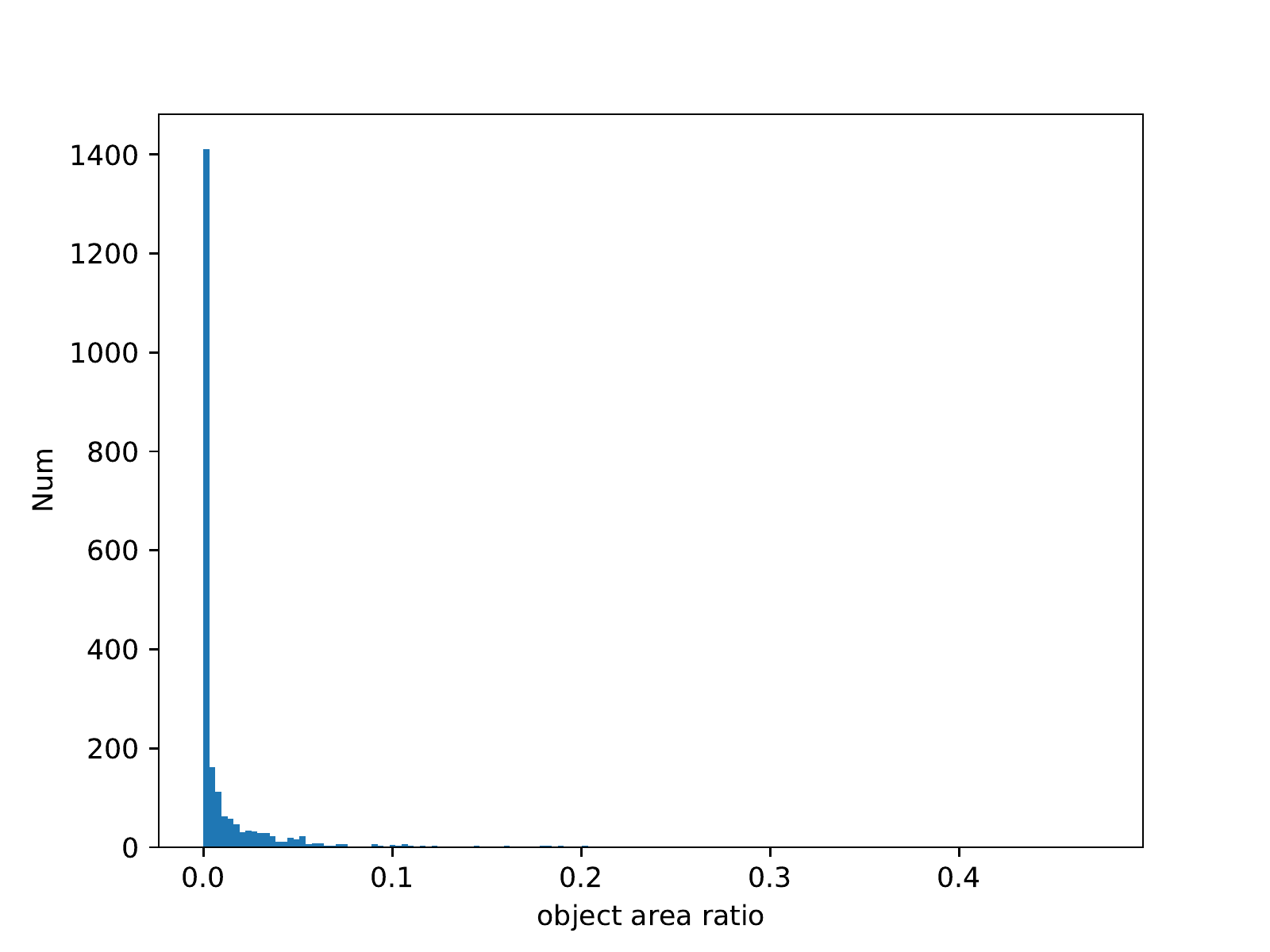}}
    \subfigure[Scale (Detection-val)]{\includegraphics[width=0.24\textwidth]{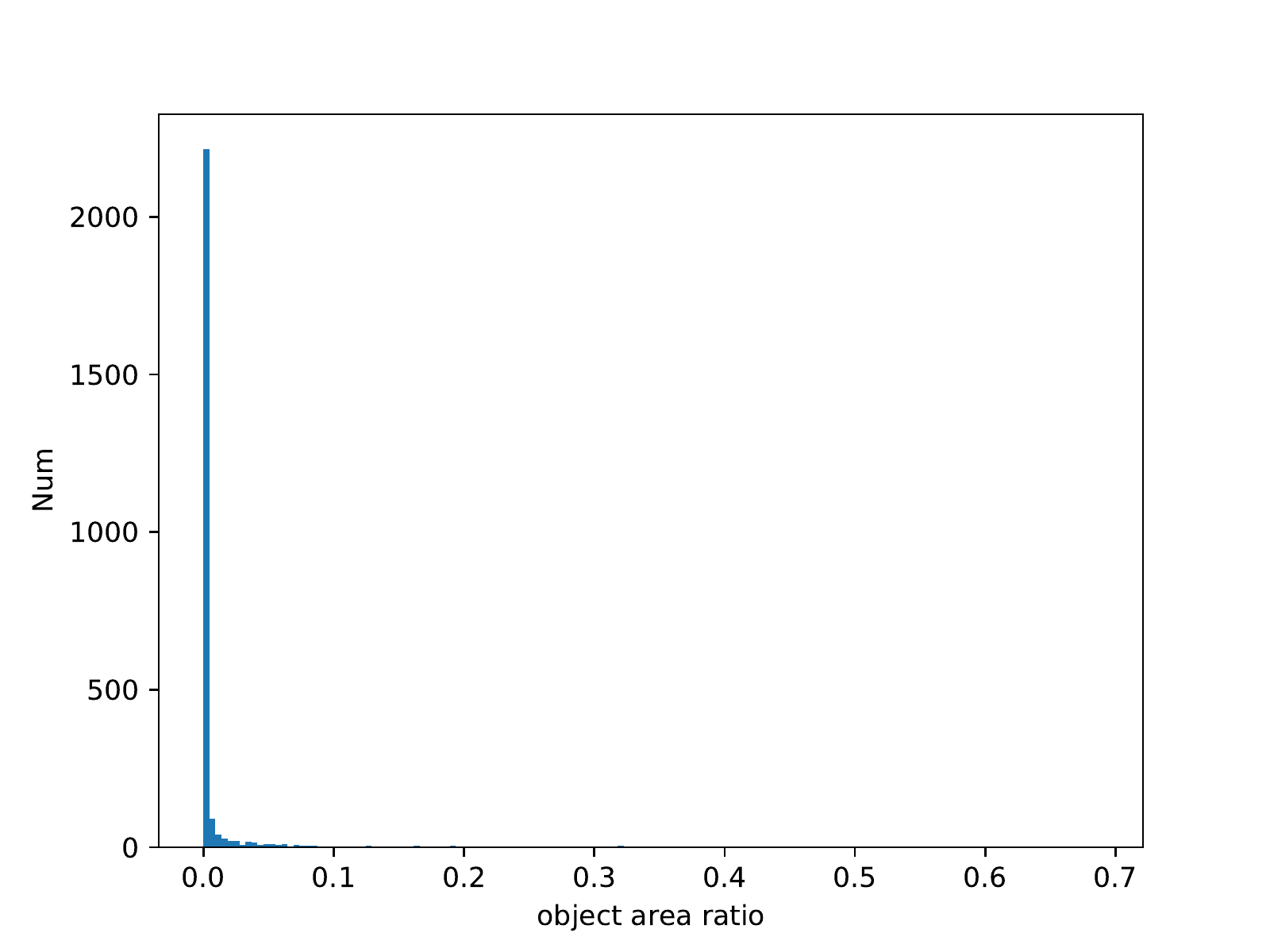}} 
    \subfigure[Scale (Tracking)]{\includegraphics[width=0.24\textwidth]{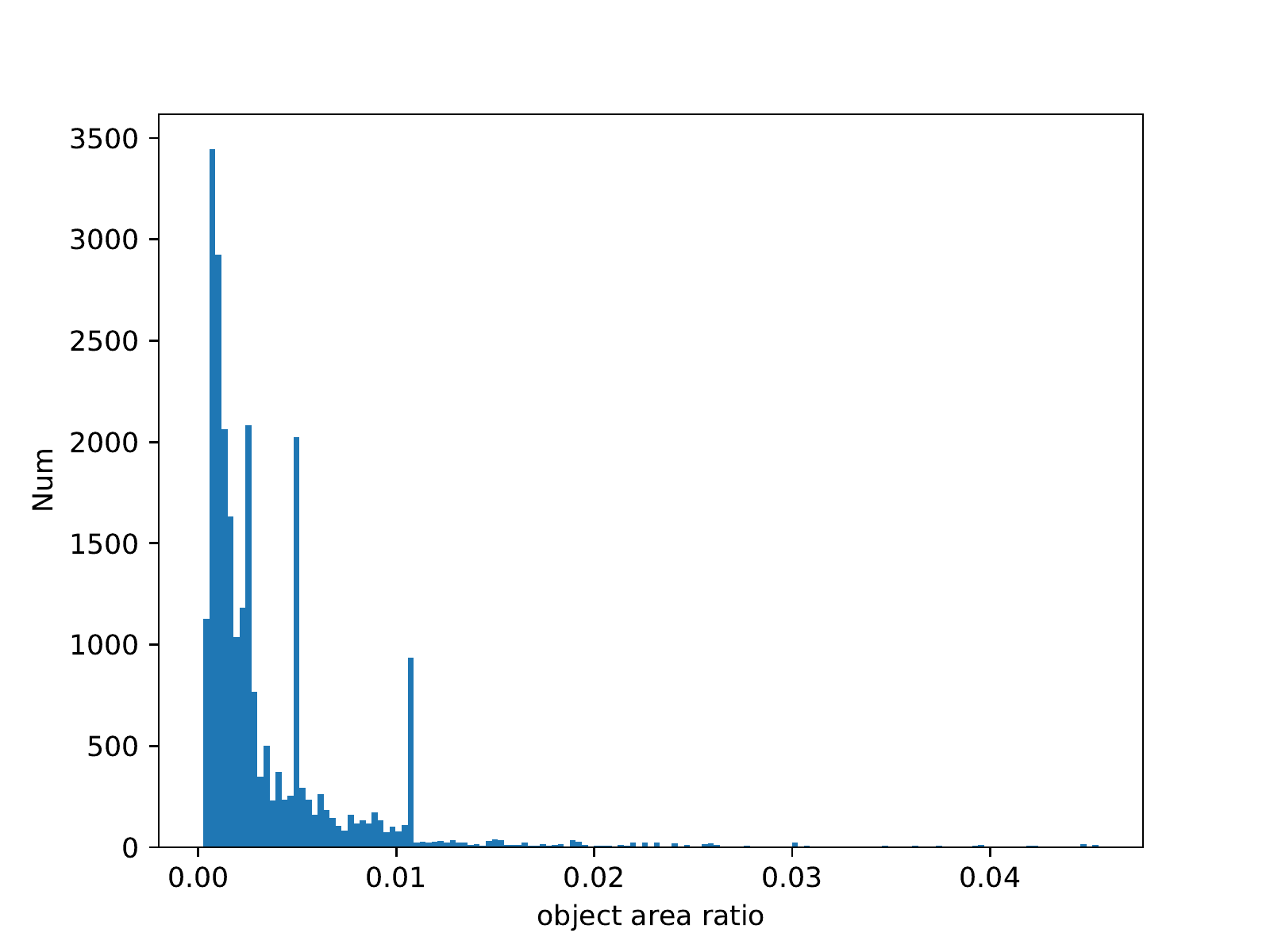}} 
    \caption{Aspect ratio and scale distribution of the DUT Anti-UAV dataset.}
    \label{fig:scale_dis}
\end{figure*}
\textbf{Anti-UAV}~\cite{jiang2021anti}. This is a dataset labeled with visible and infrared dual-mode information, which consists of 318 fully labeled videos. One hundred and sixty of the videos are used as the training set, 91 are used as the testing set, and the rest are used as the validation set, with a total of 186,494 images. The UAVs in the dataset are divided into seven attributes, which systematically conclude several special circumstances that might appear in UAV detection tasks. The recorded videos contain two environments, namely, day and night. In the two environments, the detection of the two modals plays different roles. From the perspective of location distribution, the range of motion of Anti-UAV is wide, but mostly concentrated in the central area, and it has a smaller variance compared with the two other datasets and our dataset. This dataset focuses on solving the problem that vision-based detector would show poor performance in night, while our dataset aims to improve models' robustness through enriching the diversity of multiple aspects, such as different UAV types, diverse scene information, various light conditions, and different weathers.\par \ignore{In terms of UAV size, the  distribution is also wide, average of which in RGB mode is $125\times59$ pixels (0.40\% of image size) and $52\times29$ pixels (0.50\% of the image size), so this also belongs to small object detection.}

Brian \emph{et al}.~\cite{isaac2021unmanned} collect and integrate the aforementioned three UAV datasets (i.e., MAV-VID~\cite{rodriguez2020adaptive}, Drone-vs-Bird~\cite{coluccia2019drone}, and Anti-UAV~\cite{jiang2021anti}), and present a benchmark performance study using state of the art four object detection (Faster-RCNN~\cite{ren2015faster}, YOLOv3~\cite{redmon2018yolov3}, SSD~\cite{liu2016ssd}, and DETR~\cite{DETR}) and three tracking methods (SORT~\cite{bewley2016simple}, DeepSORT~\cite{wojke2017simple}, and Tracktor~\cite{bergmann2019tracking}). Compared with this work, we propose a new dataset for both UAV detection and tracking tasks. Besides, our experiments are more sufficient. We evaluate 14 different versions of detectors from the combination of five types of detectors and three types of backbone networks. We also present the tracking performance of 8 various trackers on our dataset.\par
There is also a challenge~\cite{zhao20212nd} in the Anti-UAV community, which has been held twice until now. This challenge encourages novel and accurate methods for multi-scale object tracking, greatly promoting the development of this task. For example, SiamSTA~\cite{huang2021siamsta}, the winner of the 2nd Anti-UAV Challenge, proposes a spatial-temporal attention based Siamese tracker, which poses spatial and temporal constraints on generating candidate proposals with local neighborhoods.\par

\section{DUT Anti-UAV Benchmark}
\begin{table*}[htbp]
\centering
\caption{Attribute details of the DUT Anti-UAV dataset.}
\label{tab:dataset_attribute}
\resizebox{\textwidth}{!}{
\begin{tabular}{|c|c|c|c|c|c|c|c|c|c|c|}
\hline
                           &         & Num   & \multicolumn{2}{c|}{Image size} & \multicolumn{3}{c|}{Object area ratio} & \multicolumn{3}{c|}{Object aspect ratio} \\ \hline
                           &         &       & max             & min           & max        & avg         & min         & max          & avg         & min         \\ \hline
\multirow{4}{*}{Detection} & train   & 5243  & $3744\times5616$       & $160\times240$       & 0.70       & 0.013       & 2.6e-05     & 5.42         & 1.91        & 1.00        \\ \cline{2-11} 
                           & test    & 2245  & $1080\times1920$       & $360\times640$       & 0.47       & 0.014       & 4.1e-05     & 5.09         & 1.92        & 1.00        \\ \cline{2-11} 
                           & val     & 2621  & $2848\times4288$       & $213\times320$       & 0.69       & 0.013       & 1.9e-06     & 6.67         & 1.91        & 1.00        \\ \cline{2-11} 
                           & All     & 10109 & $3744\times5616$       & $160\times240$       & 0.70       & 0.013       & 1.9e-06     & 6.67         & 1.91        & 1.00        \\ \hline
\multirow{21}{*}{Tracking} & video01 & 1050  & \multicolumn{2}{c|}{$1080\times1920$}  & 0.021      & 0.0048      & 0.0012      & 2.59         & 2.10        & 1.64        \\ \cline{2-11} 
                           & video02 & 83    & \multicolumn{2}{c|}{$720\times1280$}   & 0.0047     & 0.0017      & 8.5e-04     & 3.62         & 2.98        & 2.71        \\ \cline{2-11} 
                           & video03 & 100   & \multicolumn{2}{c|}{$720\times1280$}   & 0.012      & 0.0023      & 6.0e-04     & 1.18         & 1.08        & 1.00        \\ \cline{2-11} 
                           & video04 & 341   & \multicolumn{2}{c|}{$1080\times1920$}  & 0.011      & 0.0032      & 6.4e-04     & 2.28         & 1.36        & 1.00        \\ \cline{2-11} 
                           & video05 & 450   & \multicolumn{2}{c|}{$720\times1280$}   & 0.0032     & 0.0020      & 8.6e-04     & 1.50         & 1.34        & 1.04        \\ \cline{2-11} 
                           & video06 & 200   & \multicolumn{2}{c|}{$1080\times1920$}  & 0.011      & 0.0044      & 0.0011      & 1.61         & 1.28        & 1.14        \\ \cline{2-11} 
                           & video07 & 2480  & \multicolumn{2}{c|}{$720\times1280$}   & 0.045      & 0.012       & 0.0023      & 4.00         & 2.12        & 1.00        \\ \cline{2-11} 
                           & video08 & 2305  & \multicolumn{2}{c|}{$720\times1280$}   & 0.030      & 0.0056      & 0.0011      & 3.98         & 2.32        & 1.00        \\ \cline{2-11} 
                           & video09 & 2500  & \multicolumn{2}{c|}{$1080\times1920$}  & 0.0099     & 0.0041      & 0.0011      & 3.82         & 1.80        & 1.00        \\ \cline{2-11} 
                           & video10 & 2635  & \multicolumn{2}{c|}{$1080\times1920$}  & 0.0048     & 0.0028      & 0.0015      & 4.33         & 2.26        & 1.00        \\ \cline{2-11} 
                           & video11 & 1000  & \multicolumn{2}{c|}{$1080\times1920$}  & 0.015      & 0.0068      & 0.0038      & 2.47         & 1.87        & 1.00        \\ \cline{2-11} 
                           & video12 & 1485  & \multicolumn{2}{c|}{$1080\times1920$}  & 0.0056     & 0.0017      & 4.9e-04     & 2.41         & 1.64        & 1.00        \\ \cline{2-11} 
                           & video13 & 1915  & \multicolumn{2}{c|}{$1080\times1920$}  & 0.0048     & 7.9e-04     & 3.3e-04     & 2.92         & 1.76        & 1.00        \\ \cline{2-11} 
                           & video14 & 590   & \multicolumn{2}{c|}{$1080\times1920$}  & 0.0044     & 0.0017      & 7.9e-04     & 2.64         & 1.76        & 1.24        \\ \cline{2-11} 
                           & video15 & 1350  & \multicolumn{2}{c|}{$1080\times1920$}  & 0.0048     & 0.0021      & 5.2e-04     & 1.94         & 1.21        & 1.00        \\ \cline{2-11} 
                           & video16 & 1285  & \multicolumn{2}{c|}{$1080\times1920$}  & 0.0014     & 8.0e-04     & 3.6e-04     & 2.29         & 1.29        & 1.00        \\ \cline{2-11} 
                           & video17 & 780   & \multicolumn{2}{c|}{$1080\times1920$}  & 0.0032     & 0.0010      & 4.9e-04     & 2.55         & 1.43        & 1.00        \\ \cline{2-11} 
                           & video18 & 1320  & \multicolumn{2}{c|}{$1080\times1920$}  & 0.0048     & 0.0013      & 3.3e-04     & 2.47         & 1.54        & 1.00        \\ \cline{2-11} 
                           & video19 & 1300  & \multicolumn{2}{c|}{$1080\times1920$}  & 0.0018     & 7.4e-04     & 2.7e-04     & 2.52         & 1.50        & 1.00        \\ \cline{2-11} 
                           & video20 & 1635  & \multicolumn{2}{c|}{$1080\times1920$}  & 0.0035     & 0.0016      & 5.8e-04     & 2.76         & 1.75        & 1.00        \\ \cline{2-11} 
                           & All     & 24804 & \multicolumn{2}{c|}{-}          & 0.045      & 0.0031      & 2.7e-04     & 4.33         & 1.72        & 1.00        \\ \hline 
\end{tabular}}
\end{table*}
\begin{figure}
    \centering
    \includegraphics[width=0.5\textwidth]{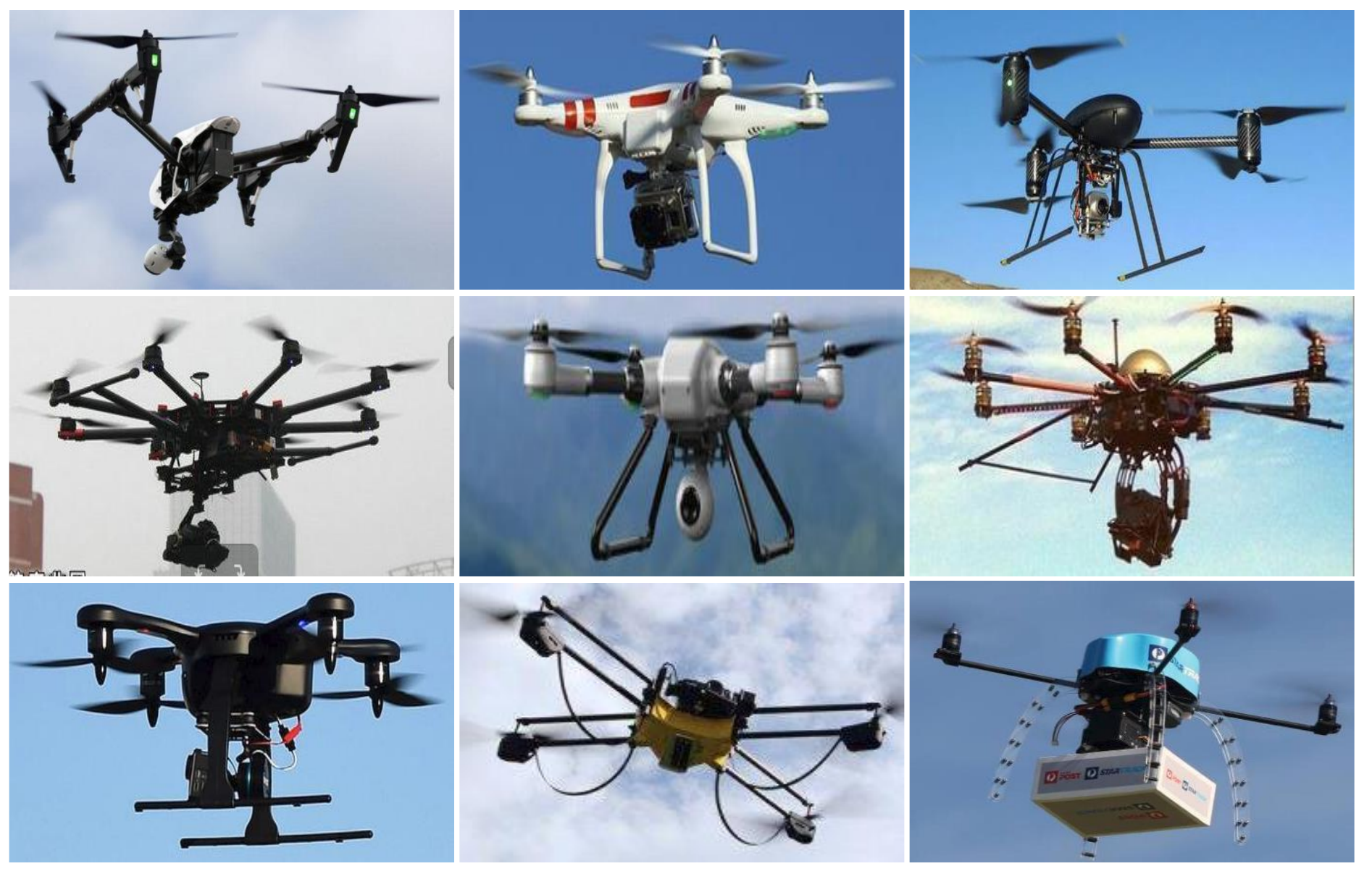}
    \caption{Examples of different types of UAVs in our dataset.}
    \label{fig:uavType}
\end{figure}
To assist in the development of the area of UAV detection and tracking, we propose a UAV detection and tracking dataset, named DUT Anti-UAV. It contains detection and tracking subsets. The detection dataset is split into three sets, namely, training, testing, and verification sets. The tracking dataset contains 20 sequences where the targets are various UAVs. It is used to test the performance of algorithms for UAV tracking.\par

\subsection{Dataset splitting}
Our DUT Anti-UAV dataset contains detection and tracking subsets. The detection dataset is split into training, testing, and validation sets. The tracking dataset contains 20 short-term and long-term sequences. All frames and images are manually annotated precisely. The detailed information of images and objects is shown in Table~\ref{tab:dataset_attribute}. Specifically, the detection dataset contains 10,000 images in total, in which the training, testing, and validation sets have 5200, 2200 and 2600 images, respectively. In consideration of the situation that one image contains multiple objects, the total number of detection objects is 10,109, where the training, testing, and validation sets have 5243, 2245, and 2621 objects, respectively.\par
\begin{figure*}
    \centering
    \includegraphics[width=0.98\textwidth]{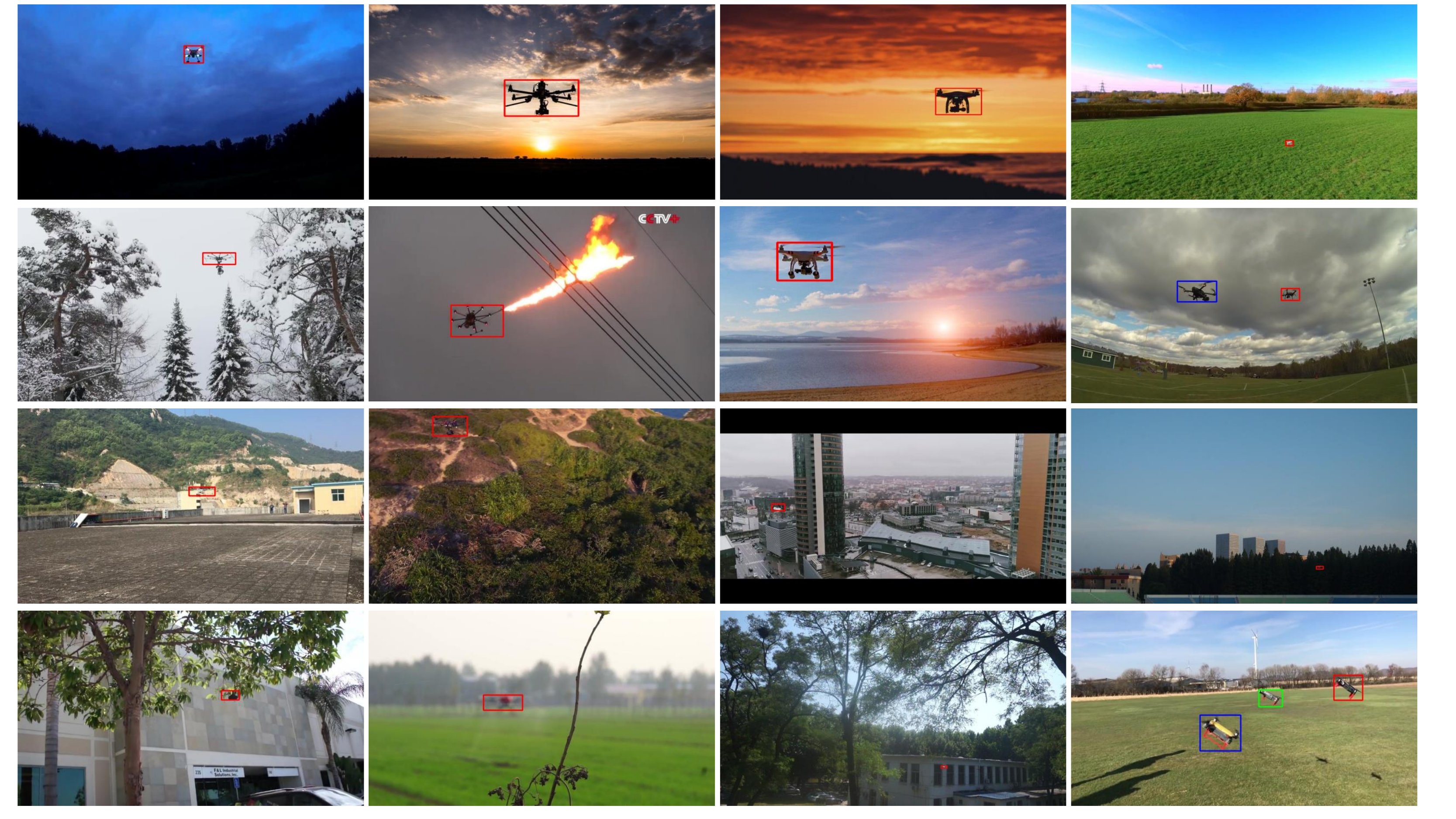}
    \caption{Examples of the detection images and annotations of our dataset.}
    \label{fig:detection_bbox}
\end{figure*}

\subsection{Dataset characteristics}
Compared with general object detection and tracking datasets (e.g., COCO~\cite{lin2014COCO}, ILSVRC~\cite{imagenet}, LaSOT~\cite{lasot}, OTB~\cite{otb2015}), the most notable characteristic of the proposed dataset for UAV detection and tracking is that the proportion of small objects is larger. In addition, given that UAVs mostly fly outdoors, the background is usually complicated, which increases the difficulty of UAV detection and tracking tasks. We analyze the characteristics of the proposed dataset from the following aspects.\par
\textbf{Image resolution}. The dataset contains images with various resolutions. For the detection dataset, the height and width of the largest image are 3744 and 5616, whereas the size of the smallest image is $160\times 240$; a huge difference between them. The tracking dataset has two type frames with $1080 \times 1920$ and $720 \times 1280$ resolutions. Various settings of image resolution can make models adapt to images with different sizes, and avoid overfitting.\par 

\textbf{Object and background}. To enrich the diversity of objects and prevent models from overfitting, we select more than 35 types of UAVs. Several examples can be seen in Fig.~\ref{fig:uavType}. The scene information in the dataset is also diverse. Given that UAVs mostly fly outdoors, the background of our dataset is an outdoor environment, including the sky, dark clouds, jungles, high-rise buildings, residential buildings, farmland, and playgrounds. Besides, a variety of light conditions (such as day, night, dawn and dusk), and different weathers (like sunny, cloudy, and snowy day) are also considered in our dataset. Various examples from the detection subset are shown in Fig.~\ref{fig:detection_bbox}. Complicated background and obvious outdoor lighting changes in our dataset are crucial for training a robust and high-performed UAV detection model.\par
\textbf{Object scale}. The sizes of UAVs are often small, and the outdoor environment is broad. Thus the proportion of small objects in our dataset is large. We calculate the object area ratio based on the full image and plot the histogram of the scale distribution, shown as Table~\ref{tab:dataset_attribute} and Fig.~\ref{fig:scale_dis}, respectively. For the detection dataset, including the training, testing, and validation sets, the average object area ratio is approximately 0.013, the smallest object area ratio is 1.9e-06, and the largest object accounts for 0.7 of the entire image. Most of the objects are small. The proportions of the objects' size in the entire image are approximately less than 0.05. For the tracking dataset, the scales of objects in the sequences change smoothly. The average object area ratio is 0.0031, the maximum ratio is 0.045, and the minimum ratio is 2.7e-04. Compared with objects in general detection and tracking datasets, small objects are much harder to detect and track, and more prone to failures, such as missed inspection and tracking loss.\par

\textbf{Object aspect ratio}. Table~\ref{tab:dataset_attribute} and Fig.~\ref{fig:scale_dis} also show the object aspect ratio. The objects in our dataset have various aspect ratios, where the maximum is 6.67, and the minimum is 1.0. In one sequence, the same object has a significant aspect ratio change. For example, the object aspect ratio in "video10" changes between 1.0 and 4.33. The aspect ratios of most objects are between 1.0 and 3.0.\par
\textbf{Object position}. Fig.~\ref{fig:pos_dis} describes the position distribution of the objects' relative center location in the form of scatter plots. Most of the objects are concentrated in the center of the image. The ranges of the object motion in all sets vary, and the horizontal and vertical movements of objects are evenly distributed. For the tracking dataset, the bounding boxes of the object in one sequence are continuous. According to Fig.~\ref{fig:pos_dis} (d), in addition to the central area of the image, objects also move frequently to the right and bottom-left of the image.\par

\subsection{Dataset challenges}
\begin{figure*}
    \centering
    \includegraphics[width=0.98\textwidth]{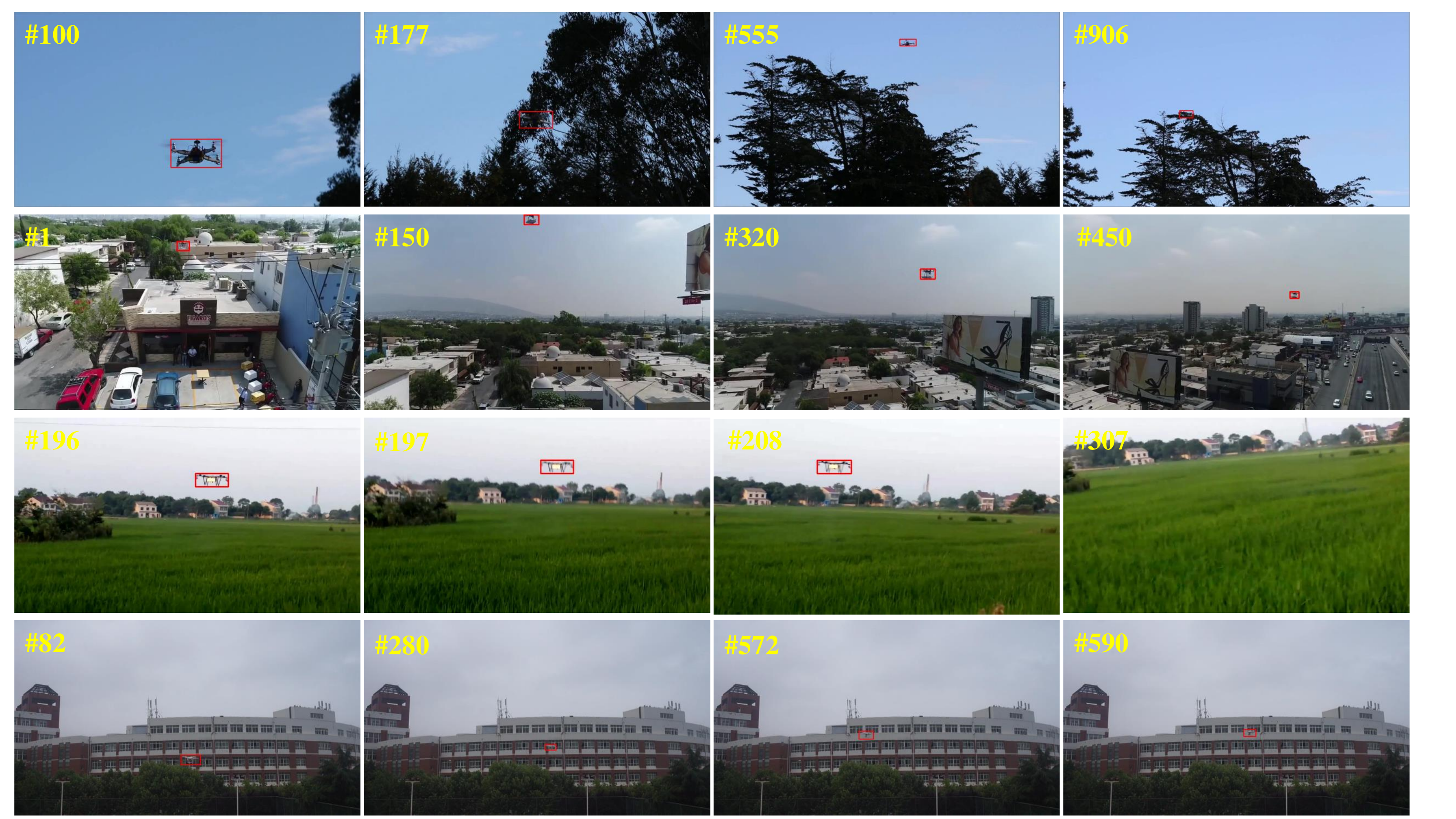}
    \caption{Examples of tracking sequences and annotations of our dataset.}
    \label{fig:tracking_bbox}
\end{figure*}

Through the analysis of the characteristics of the proposed dataset in the last subsection, we find that UAV detection and tracking encounter many difficulties and challenges. The main challenges are that the object is too small, the background is complex or similar to the object, and the light changes obviously. Object blur, fast motion, camera motion, and out of view are also prone to occur. Fig.~\ref{fig:detection_bbox} and Fig.~\ref{fig:tracking_bbox} respectively show examples of the detection and tracking datasets that reflect the aforementioned challenges.\par
\begin{table}[tbp]
\centering
\caption{Detection results of different combinations of models and backbones. The top results of mAP and FPS are marks as \bfseries{\textcolor{red}{red}}.}
\label{tab:detector_results}
\begin{tabular}{|c|c|c|c|}
\hline
                                &           & mAP                          & FPS (tasks/s)                \\ \hline
                                & ResNet50 & 0.653                        & 12.8                        \\ \cline{2-4} 
                                & ResNet18 & 0.605                        & 19.4                        \\ \cline{2-4} 
\multirow{-3}{*}{Faster-RCNN}  & VGG16    & 0.633                        & 9.3                         \\ \hline
                                & ResNet50 & \bfseries{{\color[HTML]{FE0000} 0.683}} & 10.7                        \\ \cline{2-4} 
                                & ResNet18 & 0.652                        & 14.7                        \\ \cline{2-4} 
\multirow{-3}{*}{Cascade-RCNN} & VGG16    & 0.667                        & 8.0                         \\ \hline
                                & ResNet50 & 0.642                        & 13.3                        \\ \cline{2-4} 
                                & ResNet18 & 0.61                         & 20.5                        \\ \cline{2-4} 
\multirow{-3}{*}{ATSS}          & VGG16    & 0.641                        & 9.5                         \\ \hline
                                & ResNet50 & 0.427                        & 21.7                        \\ \cline{2-4} 
                                & ResNet18 & 0.400                        & \bfseries{{\color[HTML]{FE0000} 53.7}}                        \\ \cline{2-4} 
                                & VGG16    & 0.551                        & 23.0 \\ \cline{2-4} 
\multirow{-4}{*}{YOLOX}        & DarkNet   & 0.552                        & 51.3                        \\ \hline
SSD                             & VGG16    & 0.632                        & 33.2                        \\ \hline
\end{tabular}
\end{table}
\section{Experiments}

\begin{figure*}[htbp]
    \centering
    \subfigure[IoU=0.5]{\includegraphics[width=0.45\textwidth]{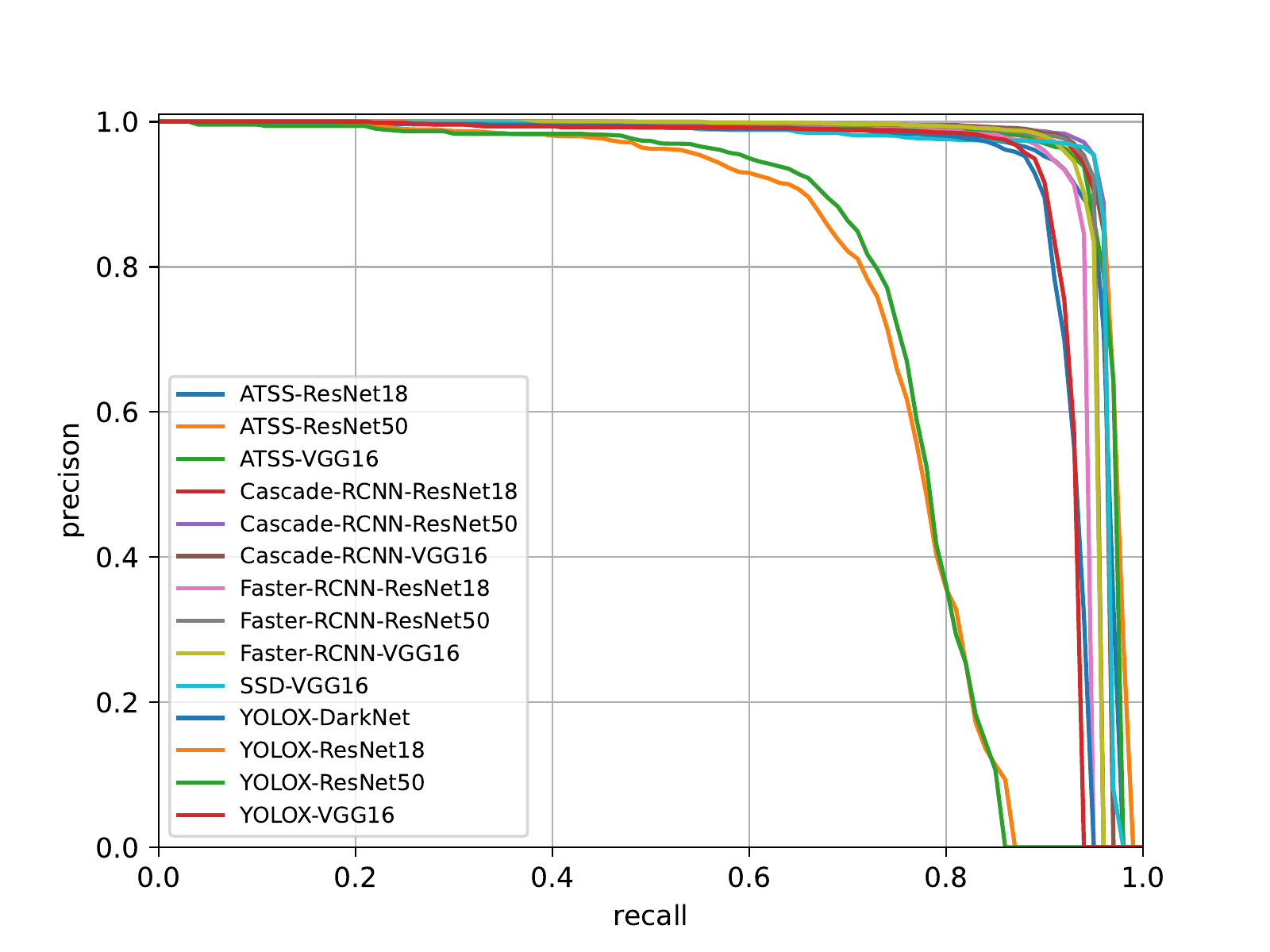}} \&
    \subfigure[IoU=0.75]{\includegraphics[width=0.45\textwidth]{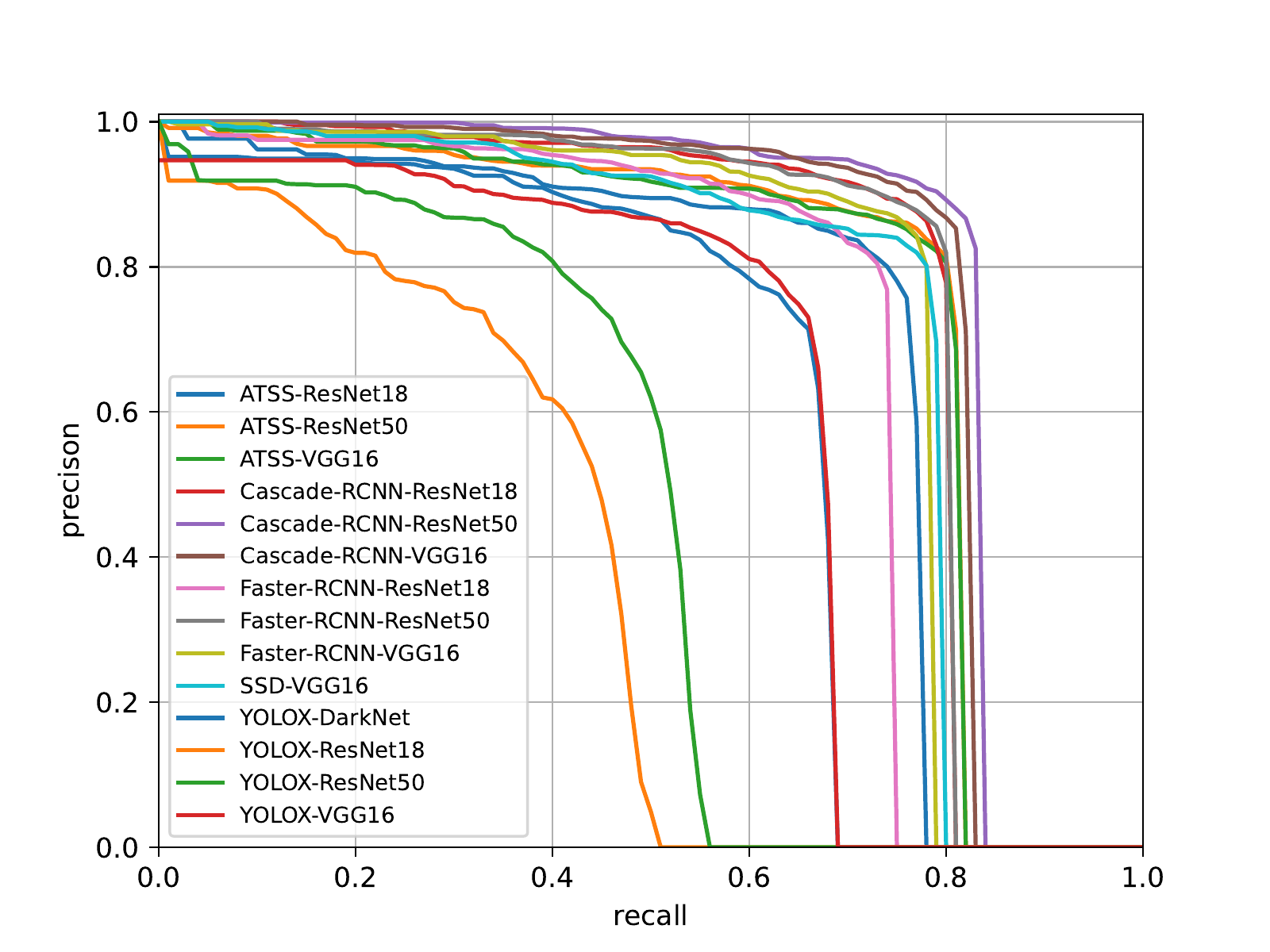}} \&
    \caption{P-R curves of all detectors.}
    \label{fig:P-R}
\end{figure*}
\subsection{Detection on DUT Anti-UAV dataset}
We select several state-of-the-art detection methods. We use Faster-RCNN~\cite{ren2015faster}, Cascade-RCNN~\cite{cai2018cascade}, and ATSS~\cite{zhang2020bridging}, which are two-stage methods, and YOLOX~\cite{8ge2021yolox} and SSD~\cite{liu2016ssd}, which are one-stage methods. Two-stage models typically have higher accuracy, while one-stage models perform better in terms of the speed. Descriptions of these algorithms are provided below.\par
\textbf{Faster-RCNN~\cite{ren2015faster}}. This method makes several improvements to Fast-RCNN~\cite{girshick2015fast} by resolving time-consuming issues on region proposals brought by selective search. Instead of selective search, the region proposal network (RPN) is proposed. This network has two branches, namely, classification and regression. Classification and regression are performed twice, so the precision of the method is high.\par
\textbf{Cascade-RCNN~\cite{cai2018cascade}}. It consists of a series of detectors with increasing Intersection over Union (IoU) thresholds. The detectors are trained stage by stage, and the output of a detector is the input of the next in which the IoU threshold is higher (in other words, a detector with higher quality). This method guarantees the amount of every detector, thereby reducing the overfitting problem.\par
\textbf{ATSS~\cite{zhang2020bridging}}. It claims that the essential difference between anchor-based and anchor-free detectors is the way of defining positive and negative training samples. It proposes an algorithm that can select positive and negative samples according to the object’s statistical feature.\par 

\textbf{YOLO~\cite{8ge2021yolox}}. YOLO series is known for its extremely high speed and relatively high accuracy. With the development of object detection, it can integrate most advanced technologies, so as to achieve rounds of iteration. After YOLOv5 reaches a peak performance, YOLOX~\cite{8ge2021yolox} starts to focus on anchor-free detectors, advanced label assignment strategies, and end-to-end (NMS-free) detectors, which are major advances in these years. After upgrading, it shows remarkable performance compared to YOLOv3~\cite{redmon2018yolov3} on COCO (a detection dataset named Common Objects in Context)~\cite{lin2014COCO}.\par
\textbf{SSD~\cite{liu2016ssd}}. It is also a one-stage detector. It combines several feature maps with different resolutions, thus improving the model’s performance via multi-scale training. It has a good effect on the detection of objects with different sizes. Only a single network is involved, making the model easy to train. \par
We replace these detectors' backbone network with several classic backbone networks, including ResNet18~\cite{resnet}, ResNet50~\cite{resnet}, and VGG16~\cite{vgg}, and obtain 14 different versions of detection methods. The 14 detectors are all retrained on the training subset of the DUT Anti-UAV detection dataset. Moreover, we use mean average precision (mAP) and frames per second (FPS) to evaluate the methods' performance. The results are shown in Table~\ref{tab:detector_results}. Cascade-RCNN with ResNet50 performs the best, and YOLOX with ResNet18 is the fastest.\par
\begin{figure*}
    \centering
    \includegraphics[width=0.98\textwidth]{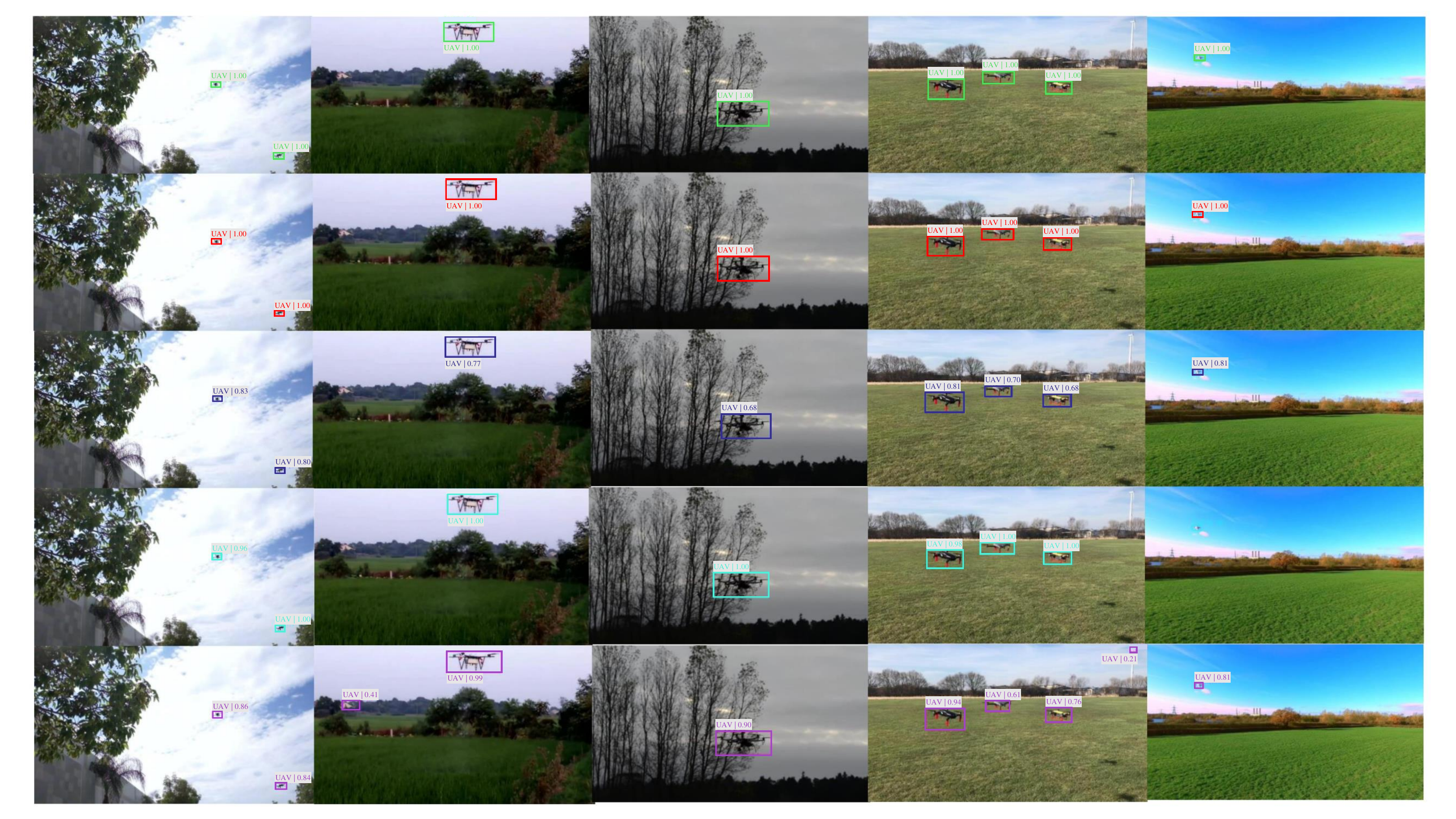}
    \caption{Qualitative comparison of detection results. The first row to the last row indicate the detection results of Faster-RCNN-ResNet50, Cascade-RCNN-ResNet50, ATSS-ResNet50, SSD-VGG16 and YOLO-DarkNet in turn, including the target bounding box and the corresponding confidence score. Better viewed with zoom-in.}
    \label{fig:bbox_plot}
\end{figure*}

We also visualize the performance of different detectors by using P-R curves with different IoU thresholds, which are shown in Fig.~\ref{fig:P-R}. In P-R curves, P means precision, and R means recall. Typically, a negative correlation exists between them, and a curve drawn with R as the abscissa and P as the ordinate can effectively reflect the comprehensive performance of a detector. Moreover, we illustrate several qualitative results in Fig.~\ref{fig:bbox_plot}. Faster-RCNN and Cascade-RCNN can get accurate bounding boxes and corresponding confidence scores, while YOLO ofen detects the background as the target mistakenly.\par

\subsection{Tracking on DUT Anti-UAV dataset}
We select several existing state-of-the-art trackers and perform them on our tracking dataset. The tracking performance is shown in the third column of Table~\ref{tab:tracker_detector_results} (the "noDET" column). We use three metric methods to evaluate the tracking performance. First, the Success calculates the IoU between the target ground truth and the predicted bounding boxes. It can reflect the accuracy of the size and scale of the predicted target bounding boxes. Second, the Precision measures the center location error by computing the distance of pixels between the ground truth and tracking results. However, it is easily affected by target size and image resolution. To solve this problem, the Norm Pre is introduced to rank trackers using Area Under Curve (AUC) between 0 and 0.5.\par
We select seven tracking algorithms in total. Their descriptions are provided below.\par
\textbf{SiamFC~\cite{SiamFC}}. This method is a classic generative tracking algorithm based on a fully convolutional Siamese network. It performs cross-correlation operation between the template patch and the search region to locate the target. Besides, a multi-scale strategy is used to decide the scale of the target.\par
\textbf{SiamRPN++~\cite{siamrpn++}}. It introduces the region proposal network (RPN) into the Siamese network, and its backbone network can be very deep. The framework has two branches, including a classification branch to choose the best anchor and a regression branch to predict offsets of the anchor. Compared with SiamFC, SiamRPN++ is more robust and faster because of the introduction of the RPN mechanism and the removal of the multi-scale strategy.\par
\textbf{ECO~\cite{eco}}. This method is a classic tracking algorithm based on correlation filtering. It introduces a factorized convolution operator to reduce model parameters. A compact generative model of the training sample space is proposed to reduce the number of training samples while guaranteeing the diversity of the sample set. Besides, an efficient model update strategy is also proposed to improve the tracker's speed and robustness.\par
\textbf{ATOM~\cite{ATOM}}. The model combines target classification and bounding box prediction. The former module is trained online to guarantee a strong discriminative capability. The latter module uses IoU loss and predicts the overlap between the target and the predicted bounding box through offline training. This combination endows the tracker with high discriminative power and good regression capability.\par
\textbf{DiMP~\cite{dimp}}. Based on ATOM, this method introduces a discriminative learning loss to guide the network to learn more discriminative features. An efficient optimizer is also designed to accelerate the convergence of the network, which improves the performance of the algorithm further.\par
\textbf{TransT~\cite{chen2021transt}}. TransT is a Transformer-based method. Due to its attention-based feature fusion network, this method is able to extract abundant semantic feature maps, and achieves state-of-the-art performance on most tracking benchmarks.\par
\textbf{SPLT~\cite{SPLT}}. It is a long-term tracker mainly based on two modules, namely, the perusal module and skimming module. The perusal module contains an efficient bounding box regressor to generate a series of target proposals, and a target verifier is used to select the best one based on confidence scores. The skimming module is used to justify the state of the target in the current frame and select an appropriate searching way (global search or local search). These improve the speed of the method, making it track in real-time.\par
\textbf{LTMU~\cite{ltmu}}. It is also a long-term tracker. The main contribution of the method is to propose a meta-updater trained offline, which is used to justify whether the tracker needs to update in the current frame or not. It greatly improves the robustness of the tracker. Moreover, a long-term tracking framework is designed based on a SiamRPN-based re-detector, an online verifier, and an online local tracker with the proposed meta-updater. This method shows its strong discriminative capability and robustness on both long-term and short-term tracking benchmarks.\par
We can find that LTMU performs the best on our tracking dataset, where the Success is 0.608 and Norm Pre is 0.783. TransT, DiMP and ATOM also exhibit well performance, with 0.586, 0.578 and 0.574 in terms of Success, respectively. The performance of SiamFC is the worst, in which the Success is 0.381 and Precision is 0.623.\par
\subsection{Tracking with detection}
\begin{figure}
    \centering
    \includegraphics[width=0.5\textwidth]{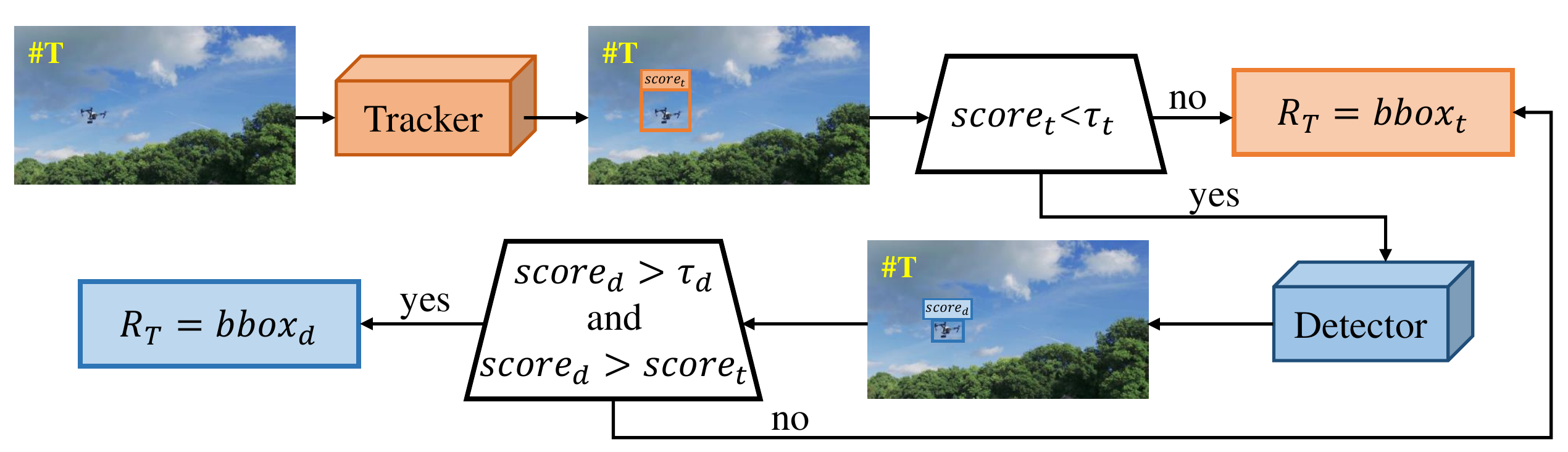}
    \caption{Framework of the proposed fusion strategy of tracking and detection.}
    \label{fig:framework}
\end{figure}

\begin{figure}
    \centering
    \includegraphics[width=0.45\textwidth]{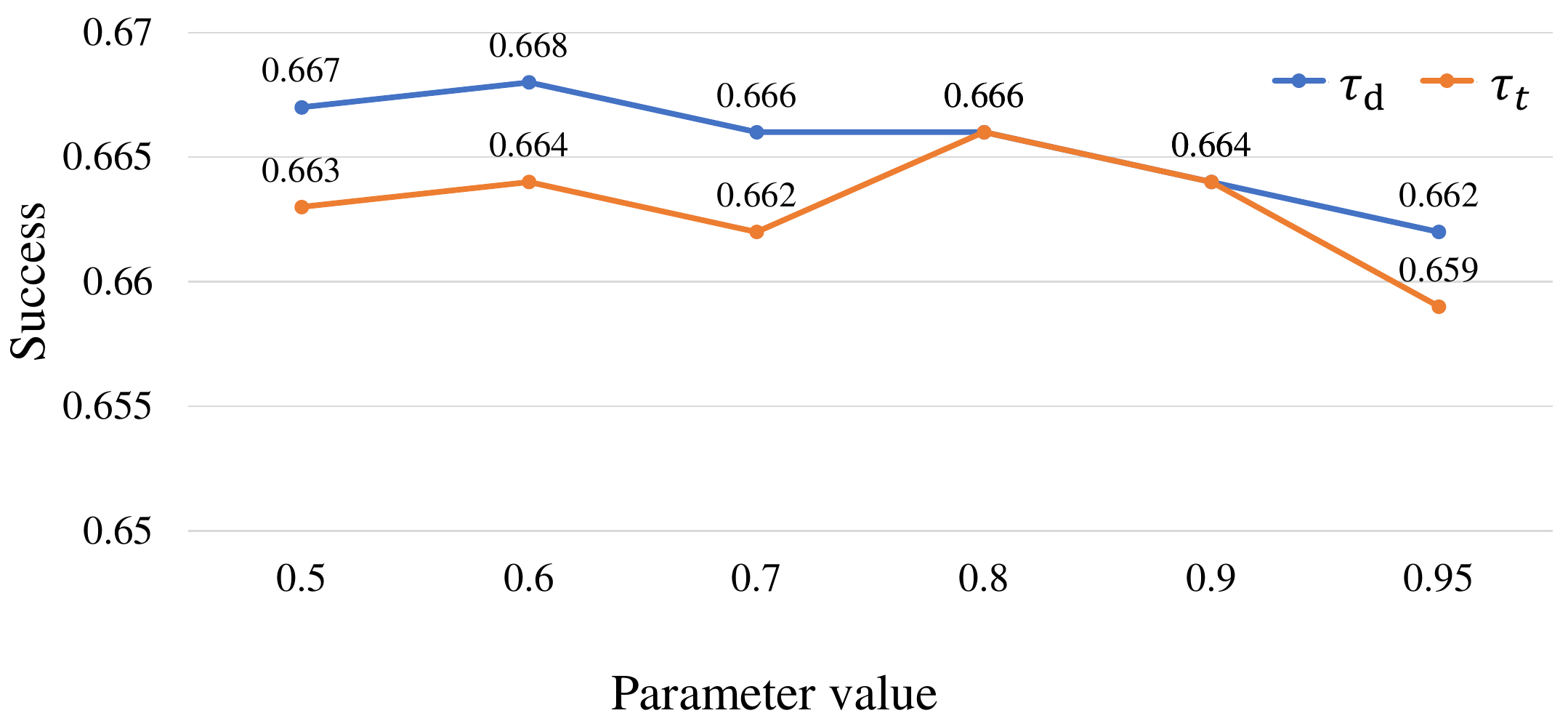}
    \caption{Effects of different parameter values for our fusion method.}
    \label{fig:ablation}
\end{figure}

\begin{table*}[htbp]
\centering
\caption{Tracking results of different combinations of trackers and detectors. The top tracking results of each tracker are marks as \textcolor{blue}{blue}, and the best results of all combinations are marked as \bfseries{\textcolor{red}{red}}.}
\label{tab:tracker_detector_results}
\resizebox{\textwidth}{!}{%
\begin{tabular}{|c|c|c|c|c|c|c|c|c|c|c|c|c|c|c|c|l|}
\hline
                            &           & noDET & \multicolumn{3}{c|}{ATSS}   & \multicolumn{3}{c|}{Cascade-RCNN}               & \multicolumn{3}{c|}{Faster-RCNN}                                                                    & SSD   & \multicolumn{4}{c|}{YOLOX}             \\ \hline
                            &           &       & ResNet18 & ResNet50 & VGG16 & ResNet18 & ResNet50                     & VGG16 & ResNet18                     & ResNet50                     & VGG16                                 & VGG16 & ResNet18 & ResNet50 & VGG16 & DarkNet \\ \hline
                            & Success   & 0.381 & 0.381    & 0.381    & 0.381 & 0.607    & {\color[HTML]{3531FF} 0.617} & 0.611 & 0.605                        & 0.616                        & 0.615                                 & 0.551 & 0.381    & 0.381    & 0.383 & 0.381   \\ \cline{2-17} 
                            & Norm Pre  & 0.526 & 0.526    & 0.526    & 0.526 & 0.792    & 0.800                        & 0.802 & 0.793                        & 0.804                        & {\color[HTML]{3531FF} 0.811}          & 0.708 & 0.526    & 0.526    & 0.528 & 0.526   \\ \cline{2-17} 
\multirow{-3}{*}{SiamFC}    & Precision & 0.623 & 0.623    & 0.623    & 0.623 & 0.920    & 0.933                        & 0.930 & 0.923                        & 0.937                        & {\color[HTML]{3531FF} 0.943}          & 0.832 & 0.623    & 0.623    & 0.625 & 0.623   \\ \hline
                            & Success   & 0.404 & 0.414    & 0.417    & 0.411 & 0.614    & 0.620                        & 0.618 & 0.610                        & 0.619                        & {\color[HTML]{3531FF} 0.620}          & 0.578 & 0.400    & 0.410    & 0.436 & 0.401   \\ \cline{2-17} 
                            & Norm Pre  & 0.643 & 0.655    & 0.663    & 0.655 & 0.809    & 0.809                        & 0.817 & 0.806                        & 0.813                        & {\color[HTML]{3531FF} 0.821}          & 0.786 & 0.643    & 0.656    & 0.687 & 0.644   \\ \cline{2-17} 
\multirow{-3}{*}{ECO}       & Precision & 0.717 & 0.717    & 0.726    & 0.717 & 0.938    & 0.942                        & 0.945 & 0.937                        & 0.946                        & {\color[HTML]{3531FF} 0.954}          & 0.884 & 0.706    & 0.718    & 0.749 & 0.705   \\ \hline
                            & Success   & 0.405 & 0.405    & 0.405    & 0.405 & 0.542    & 0.549                        & 0.546 & 0.543                        & 0.550                        & {\color[HTML]{3531FF} 0.553}          & 0.496 & 0.603    & 0.605    & 0.597 & 0.607   \\ \cline{2-17} 
                            & Norm Pre  & 0.585 & 0.585    & 0.585    & 0.585 & 0.766    & 0.775                        & 0.772 & 0.770                        & 0.779                        & {\color[HTML]{3531FF} 0.783}          & 0.700 & 0.779    & 0.780    & 0.769 & 0.783   \\ \cline{2-17} 
\multirow{-3}{*}{SPLT}      & Precision & 0.651 & 0.651    & 0.651    & 0.651 & 0.855    & 0.866                        & 0.862 & 0.860                        & 0.871                        & {\color[HTML]{3531FF} 0.875}          & 0.778 & 0.855    & 0.861    & 0.846 & 0.857   \\ \hline
                            & Success   & 0.574 & 0.532    & 0.543    & 0.565 & 0.611    & 0.621                        & 0.626 & {\color[HTML]{3531FF} 0.635} & 0.633                        & 0.632                                 & 0.601 & 0.574    & 0.536    & 0.537 & 0.547   \\ \cline{2-17} 
                            & Norm Pre  & 0.758 & 0.703    & 0.742    & 0.767 & 0.791    & 0.814                        & 0.814 & {\color[HTML]{3531FF} 0.828} & 0.820                        & 0.823                                 & 0.778 & 0.763    & 0.730    & 0.722 & 0.744   \\ \cline{2-17} 
\multirow{-3}{*}{ATOM}      & Precision & 0.830 & 0.774    & 0.809    & 0.832 & 0.895    & 0.917                        & 0.918 & {\color[HTML]{3531FF} 0.936} & 0.931                        & 0.932                                 & 0.870 & 0.836    & 0.796    & 0.794 & 0.808   \\ \hline
                            & Success   & 0.545 & 0.545    & 0.545    & 0.545 & 0.606    & 0.610                        & 0.610 & 0.606                        & 0.610                        & {\color[HTML]{3531FF} 0.612}          & 0.591 & 0.545    & 0.545    & 0.546 & 0.545   \\ \cline{2-17} 
                            & Norm Pre  & 0.709 & 0.709    & 0.709    & 0.709 & 0.788    & 0.793                        & 0.793 & 0.789                        & 0.794                        & {\color[HTML]{3531FF} 0.797}          & 0.766 & 0.709    & 0.709    & 0.710 & 0.709   \\ \cline{2-17} 
\multirow{-3}{*}{SiamRPN++} & Precision & 0.780 & 0.780    & 0.780    & 0.780 & 0.870    & 0.876                        & 0.876 & 0.873                        & 0.877                        & {\color[HTML]{3531FF} 0.881}          & 0.843 & 0.780   & 0.780    & 0.781 & 0.780   \\ \hline
                            & Success   & 0.578 & 0.589    & 0.609    & 0.600 & 0.645    & 0.637                        & 0.632 & 0.647                        & {\color[HTML]{3531FF} 0.657} & 0.654                                 & 0.627 & 0.598    & 0.589    & 0.595 & 0.583   \\ \cline{2-17} 
                            & Norm Pre  & 0.756 & 0.760    & 0.796    & 0.773 & 0.840    & 0.818                        & 0.825 & 0.845                        & {\color[HTML]{3531FF} 0.856} & 0.850                                 & 0.809 & 0.772    & 0.761    & 0.768 & 0.753   \\ \cline{2-17} 
\multirow{-3}{*}{DiMP}      & Precision & 0.831 & 0.838    & 0.873    & 0.852 & 0.931    & 0.912                        & 0.915 & 0.936                        & {\color[HTML]{3531FF} 0.949} & 0.945                                 & 0.898 & 0.851    & 0.836    & 0.845 & 0.829   \\ \hline
                            & Success   & 0.586 & 0.586    & 0.586    & 0.586 & 0.623    & {\color[HTML]{3531FF} 0.624}                       & 0.623 & 0.623                        & 0.623                        & 0.623 & 0.586 &    0.586 & 0.586   & 0.586 & 0.586 \\ \cline{2-17} 
                            & Norm Pre  & 0.765 & 0.765    & 0.765    & 0.765 & 0.807    & {\color[HTML]{3531FF} 0.808}                        & 0.808 & 0.807 & 0.808 &0.808  & 0.765 & 0.765   & 0.765  & 0.765 & 0.765  \\ \cline{2-17} 
\multirow{-3}{*}{TransT}      & Precision & 0.832 & 0.832  & 0.832  & 0.832 & 0.885    & {\color[HTML]{3531FF} 0.888} & 0.888 & 0.886 & 0.888 &0.888  & 0.832 & 0.832 & 0.832   & 0.832 &  0.832  \\ \hline
                            & Success   & 0.608 & 0.605    & 0.605    & 0.600 & 0.657    & 0.659                        & 0.658 & 0.653                        & 0.659                        & {\color[HTML]{FE0000} \textbf{0.664}} & 0.623 & 0.612    & 0.600    & 0.606 & 0.606   \\ \cline{2-17} 
                            & Norm Pre  & 0.783 & 0.782    & 0.780    & 0.768 & 0.855    & 0.856                        & 0.856 & 0.851                        & 0.858                        & {\color[HTML]{FE0000} \textbf{0.865}} & 0.803 & 0.799    & 0.772    & 0.783 & 0.781   \\ \cline{2-17} 
\multirow{-3}{*}{LTMU}      & Precision & 0.858 & 0.856    & 0.855    & 0.847 & 0.948    & 0.954                        & 0.951 & 0.946                        & 0.954                        & {\color[HTML]{FE0000} \textbf{0.961}} & 0.887 & 0.874    & 0.845    & 0.858 & 0.855   \\ \hline
\end{tabular}%
}
\end{table*}
To improve the tracking performance further, and make full use of our dataset, including the detection and tracking sets, we propose a clear and simple tracking algorithm combined with detection. The fusion strategy is shown as Fig.~\ref{fig:framework} and Algorithm~\ref{alg:combination}. Given a tracker $\mathcal{T}$ and a detector $\mathcal{D}$, we first initialize the tracker $\mathcal{T}$ based on the ground truth $GT_0$ of the first frame. For each subsequent frame, we obtain the bounding box $bbox_t$ and its confidence score $score_t$ from the tracker. If $score_t$ is less than $\tau_t$, we regard it as an unreliable result, and introduce the detection mechanism. Next, the detector obtains bounding boxes $bboxes_d$ and their confidence scores $scores_d$. If the highest score $score_d$ is higher than $\tau_d$ and $score_t$, we set the corresponding detected bounding box $bbox_d$ as the current result; otherwise, $bbox_t$ is the final result. In this paper, hyper-parameters $\tau_t$ and $\tau_d$ are set to 0.9. To investigate the effects of different parameter values for our fusion method, we change values of hyper-parameters $\tau_t$ and $\tau_d$, respectively. Fig.~\ref{fig:ablation} shows that our tracker is robust to these two parameters. That is, huge changes in parameters $\tau_t$ and $\tau_d$ only cause slight fluctuations in the tracking results (less than 1\%).\par

\begin{figure*}[htbp]
    \centering
    \subfigure[SiamFC]{\includegraphics[width=0.24\textwidth]{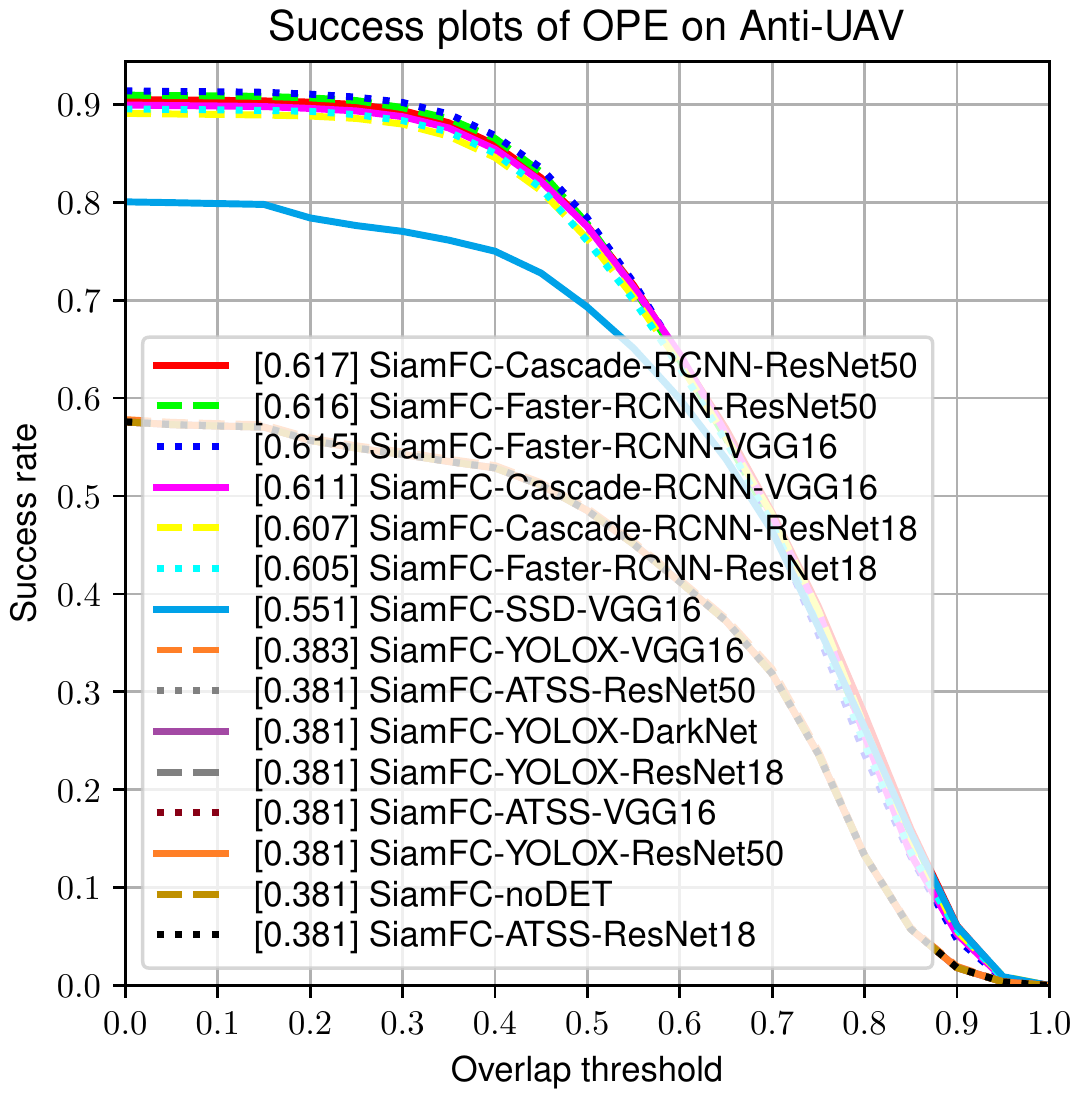}
    \includegraphics[width=0.24\textwidth]{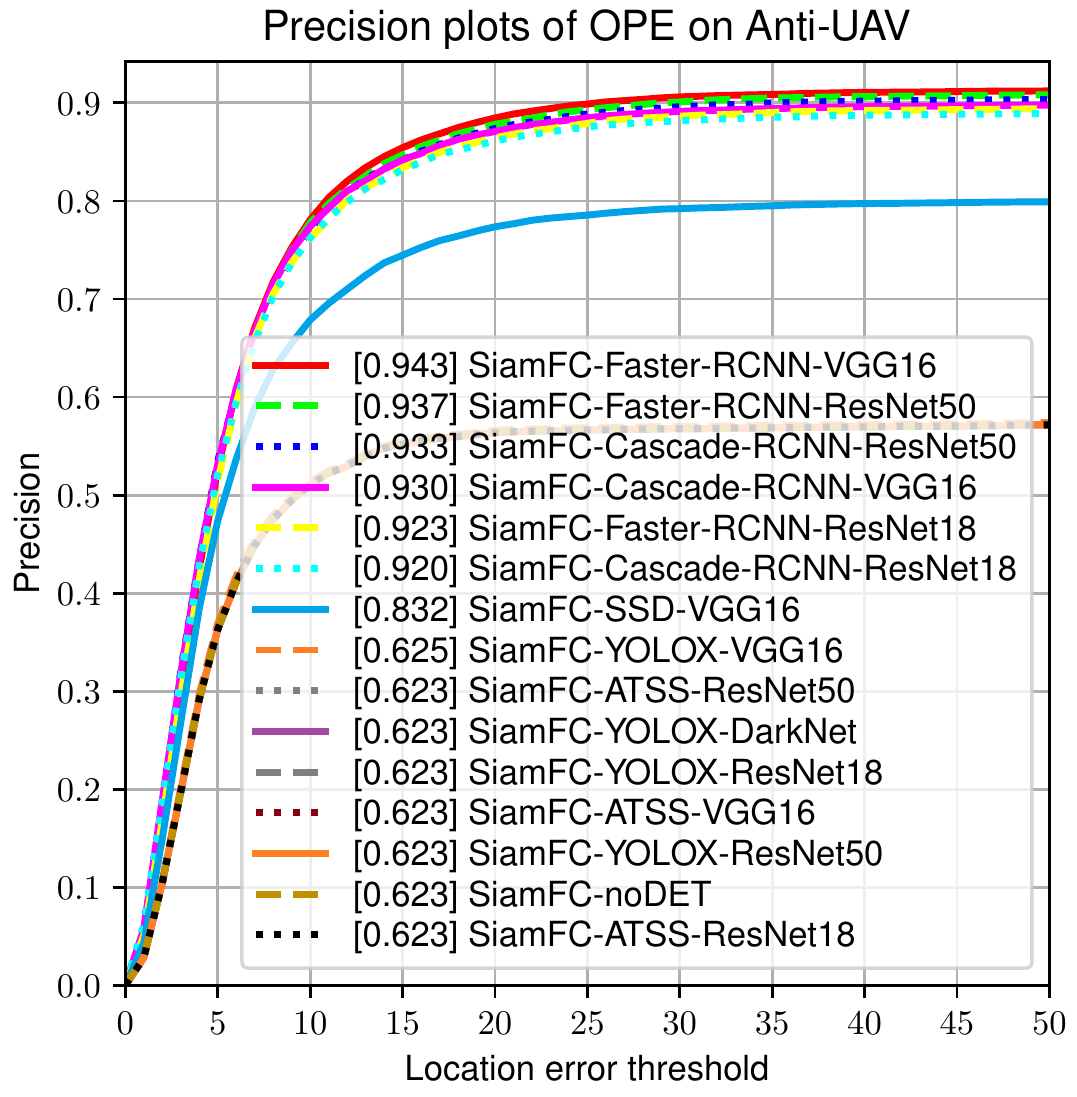}} 
    \subfigure[ECO]{\includegraphics[width=0.24\textwidth]{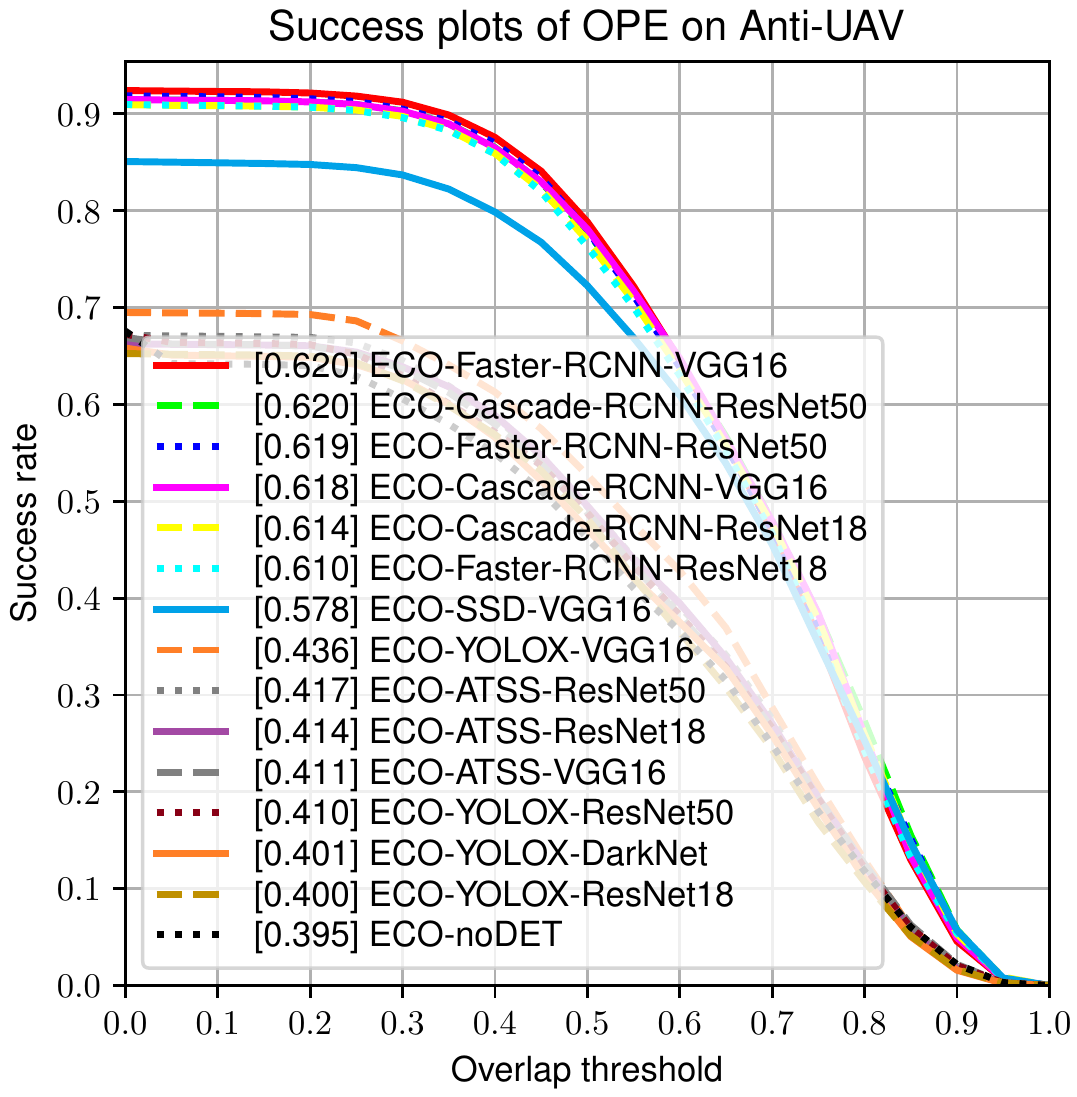}
    \includegraphics[width=0.24\textwidth]{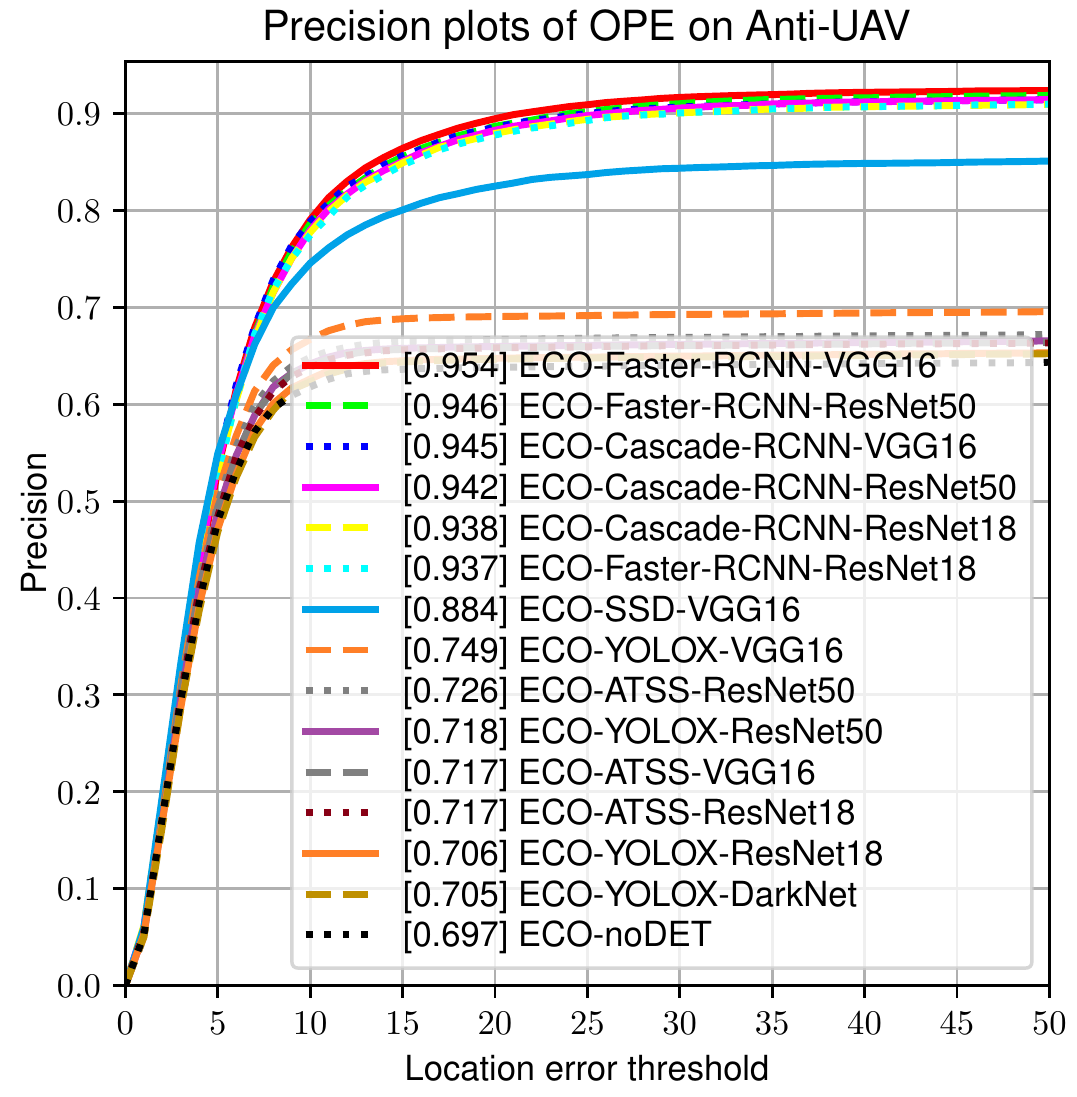}} \\
    \subfigure[SPLT]{\includegraphics[width=0.24\textwidth]{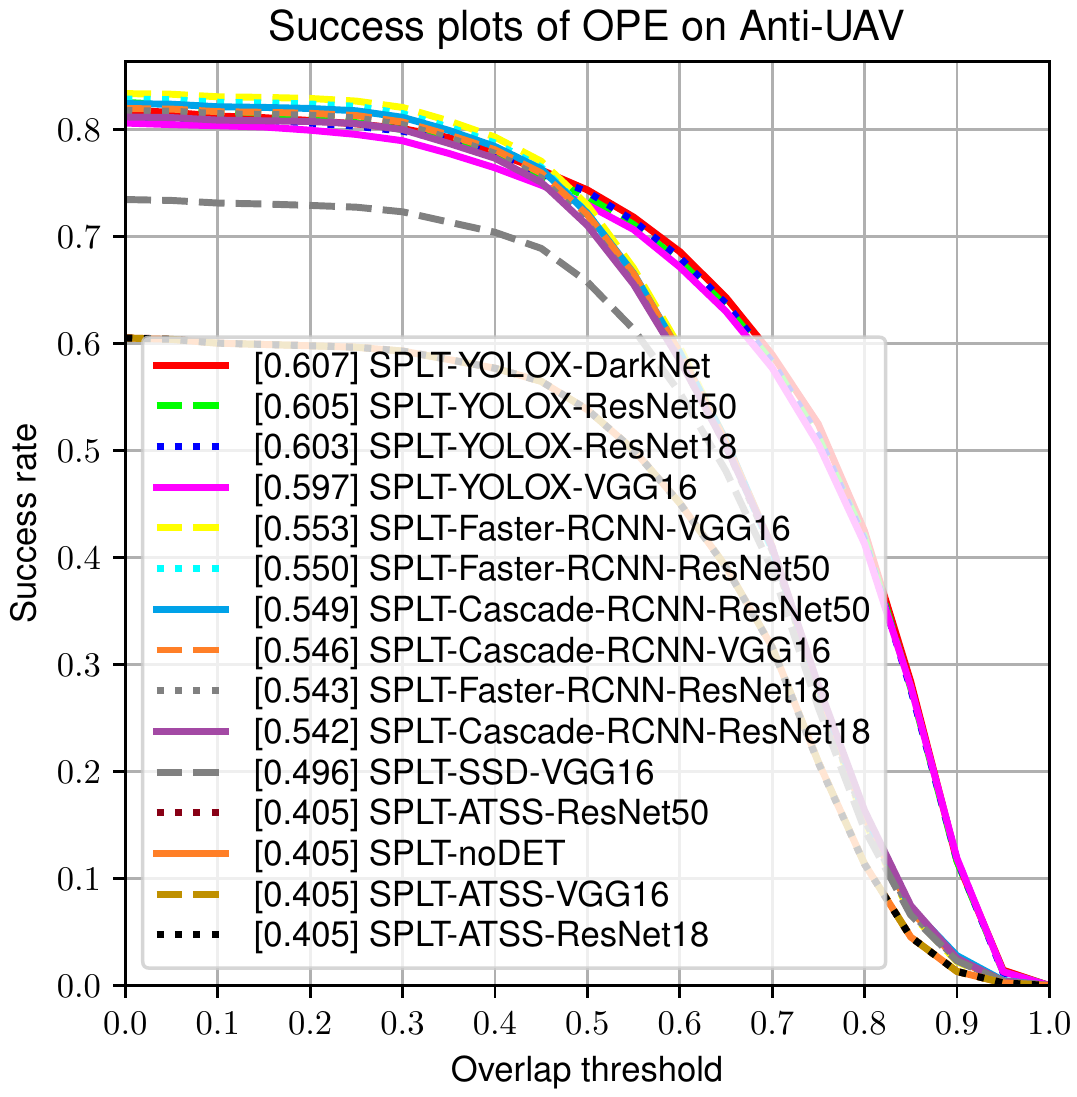}
    \includegraphics[width=0.24\textwidth]{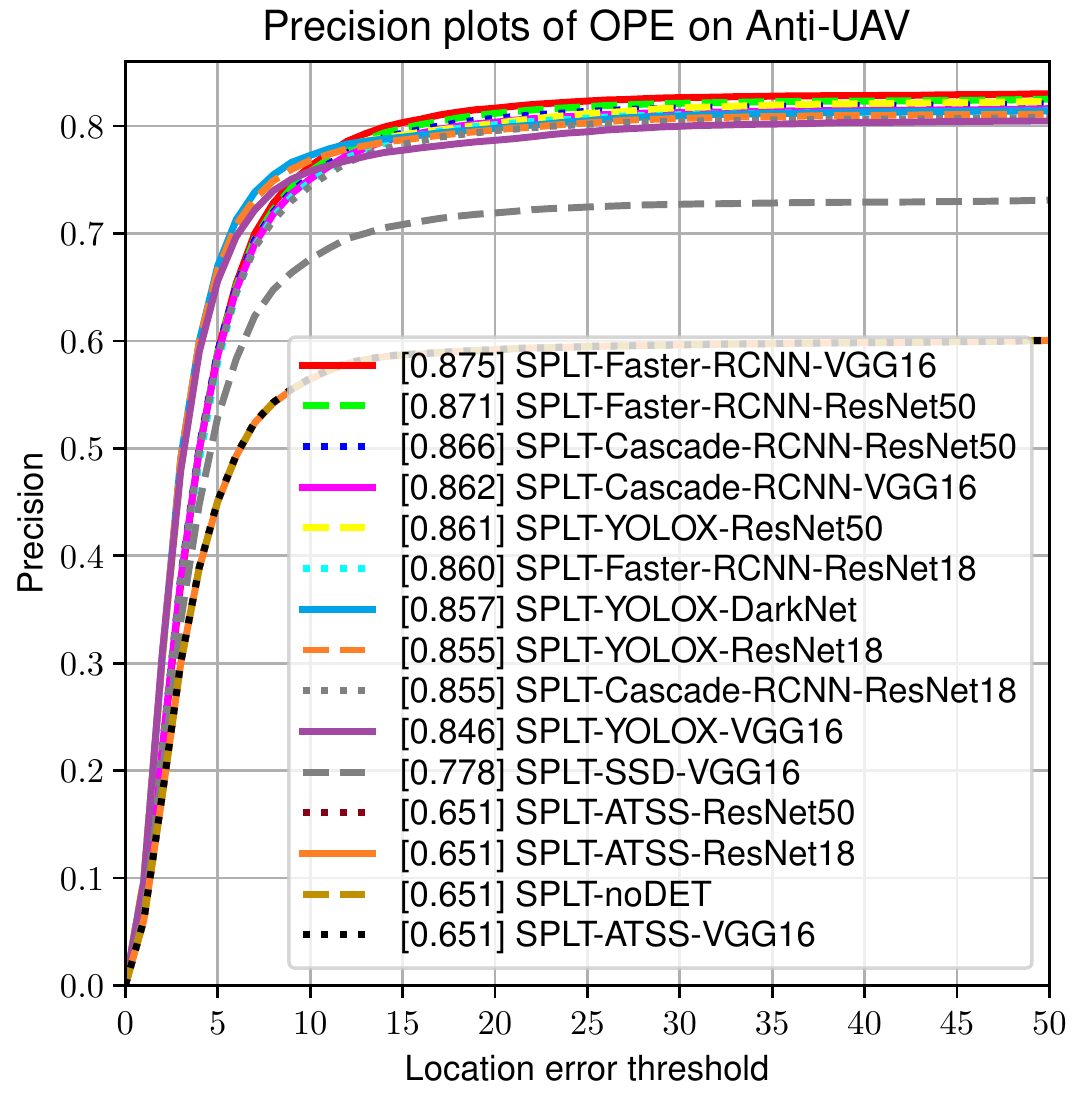}} 
    \subfigure[ATOM]{\includegraphics[width=0.24\textwidth]{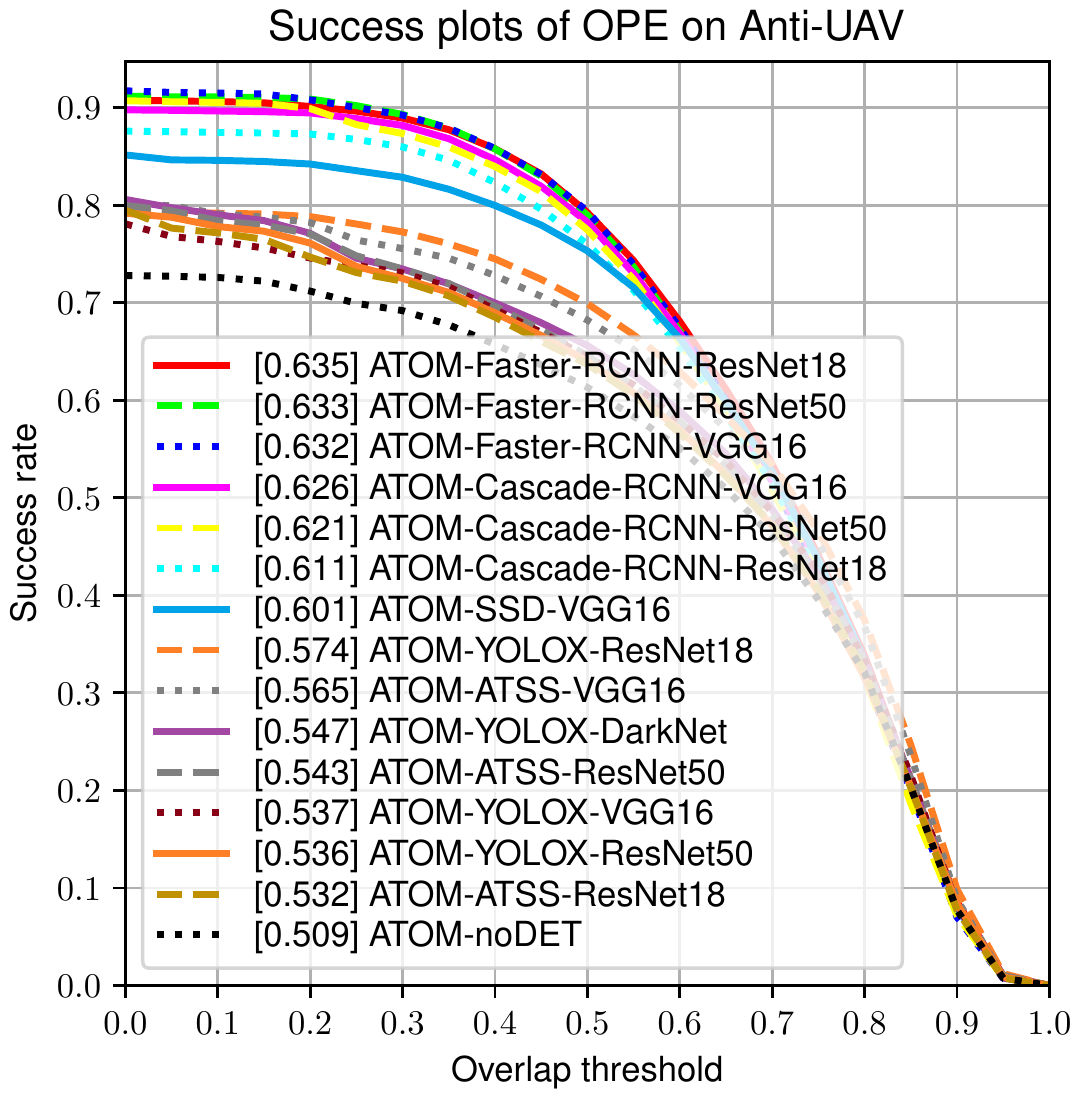}
    \includegraphics[width=0.24\textwidth]{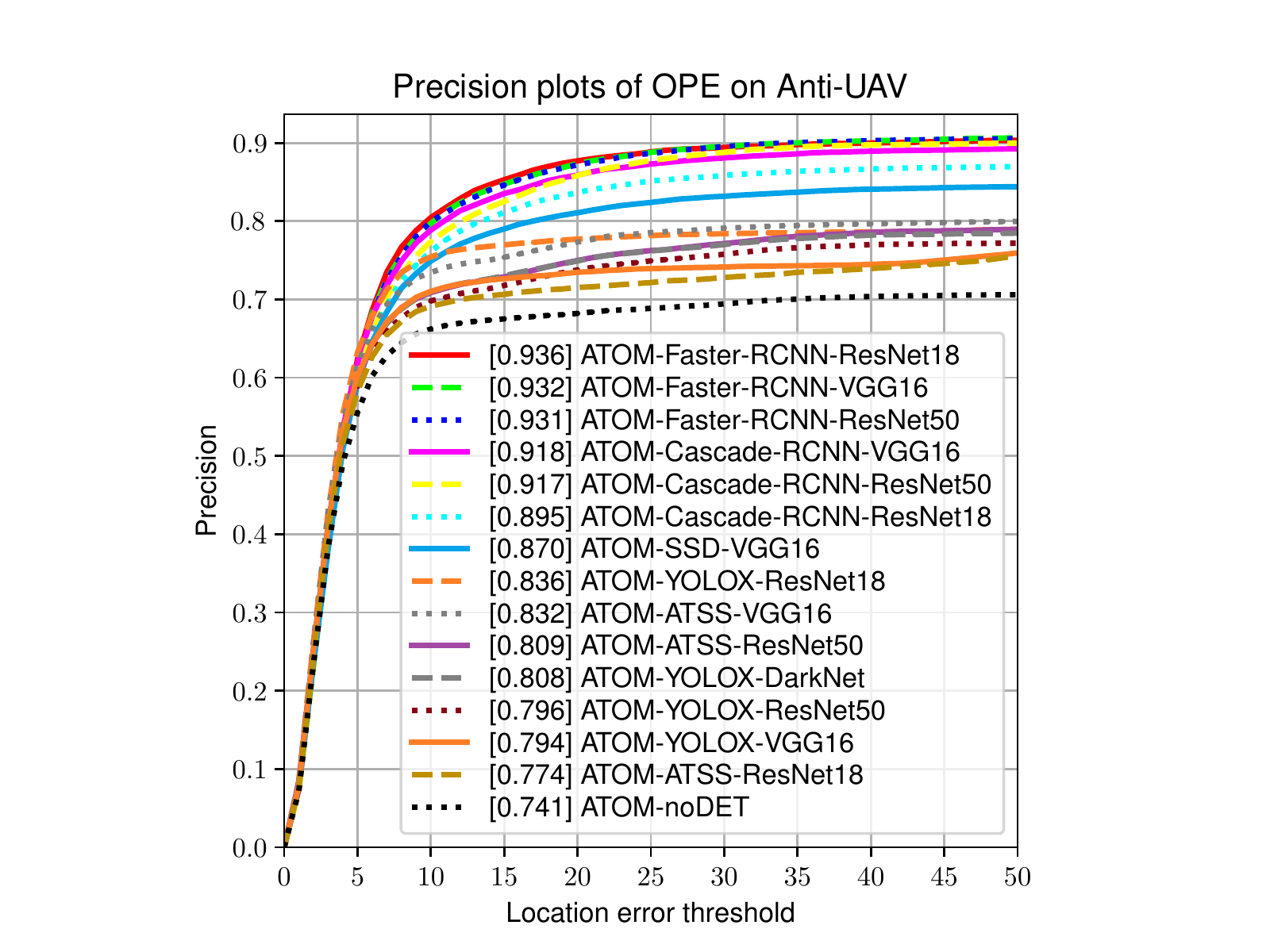}} \\
    \subfigure[SiamRPN++]{\includegraphics[width=0.24\textwidth]{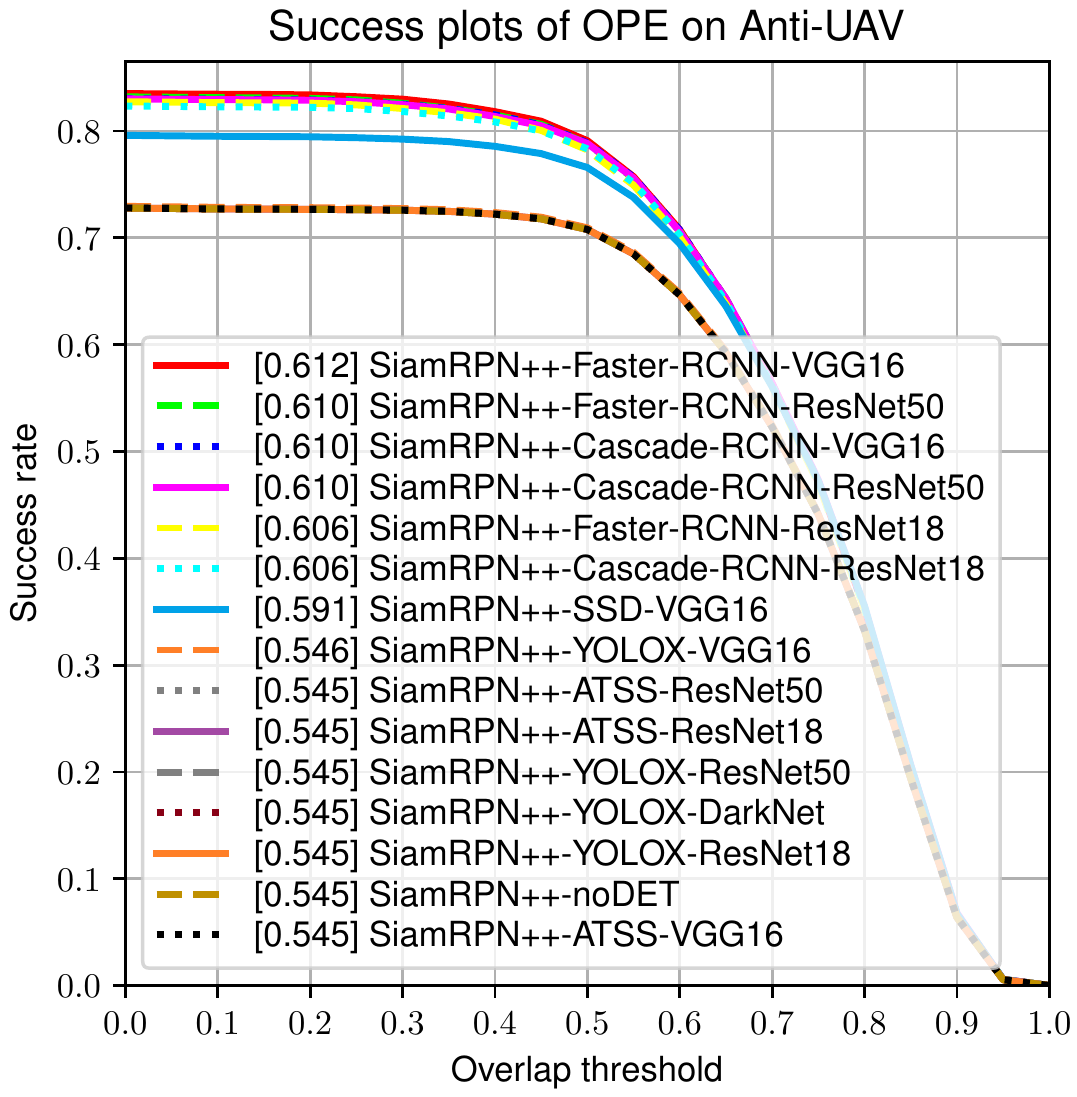}
    \includegraphics[width=0.24\textwidth]{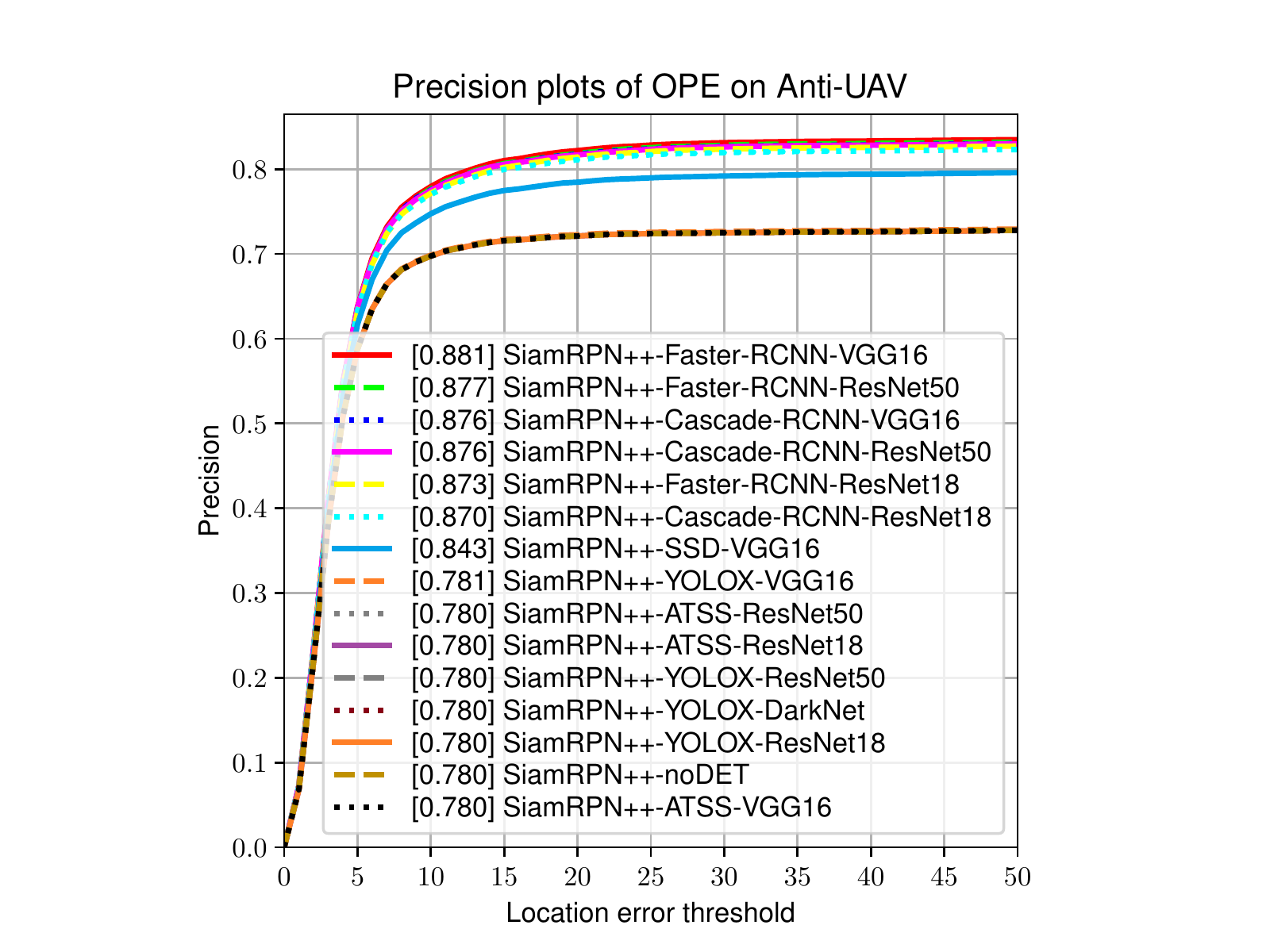}} 
    \subfigure[TransT]{\includegraphics[width=0.24\textwidth]{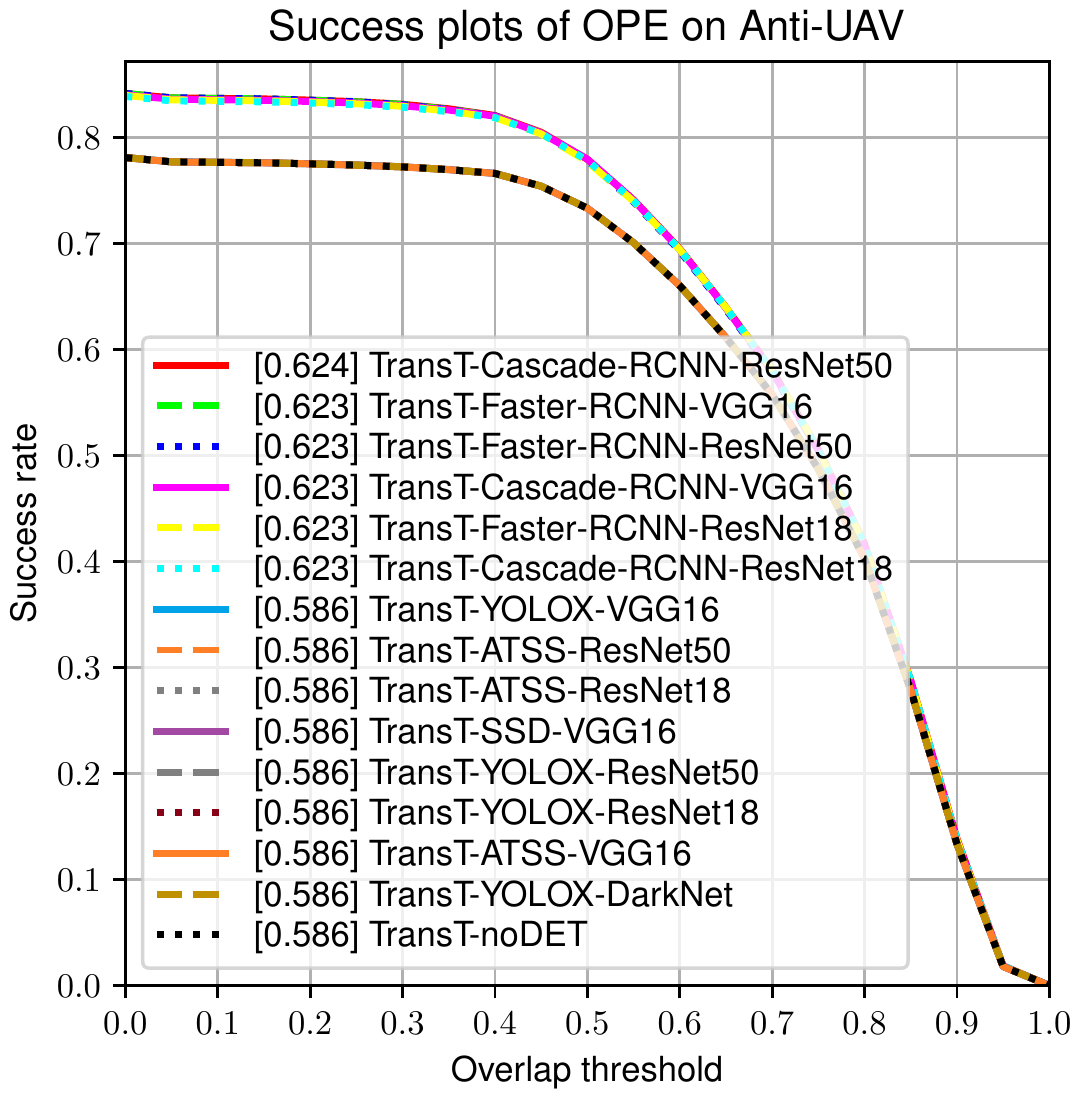}
    \includegraphics[width=0.24\textwidth]{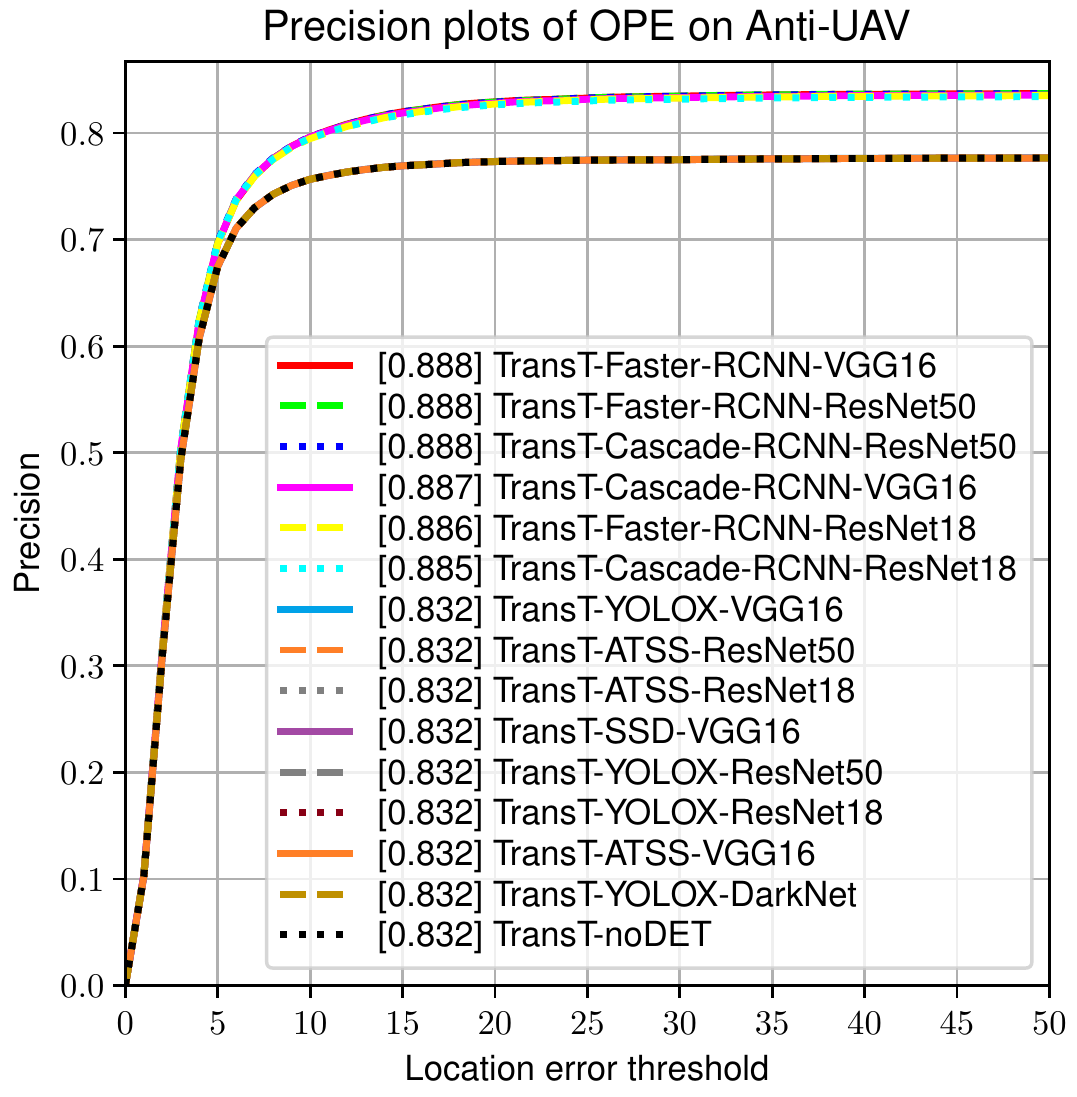}} \\
    \subfigure[DiMP]{\includegraphics[width=0.24\textwidth]{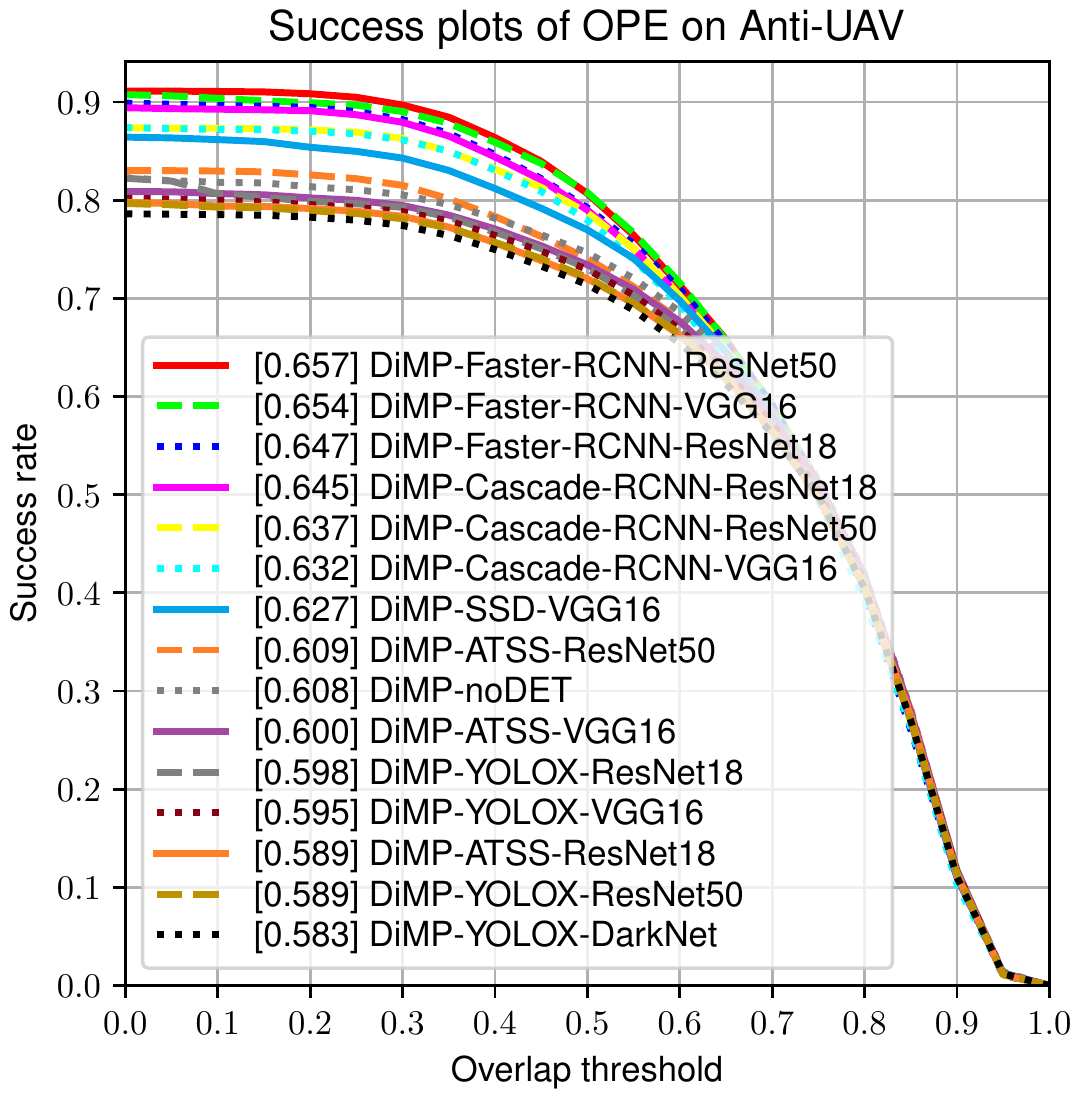}
    \includegraphics[width=0.24\textwidth]{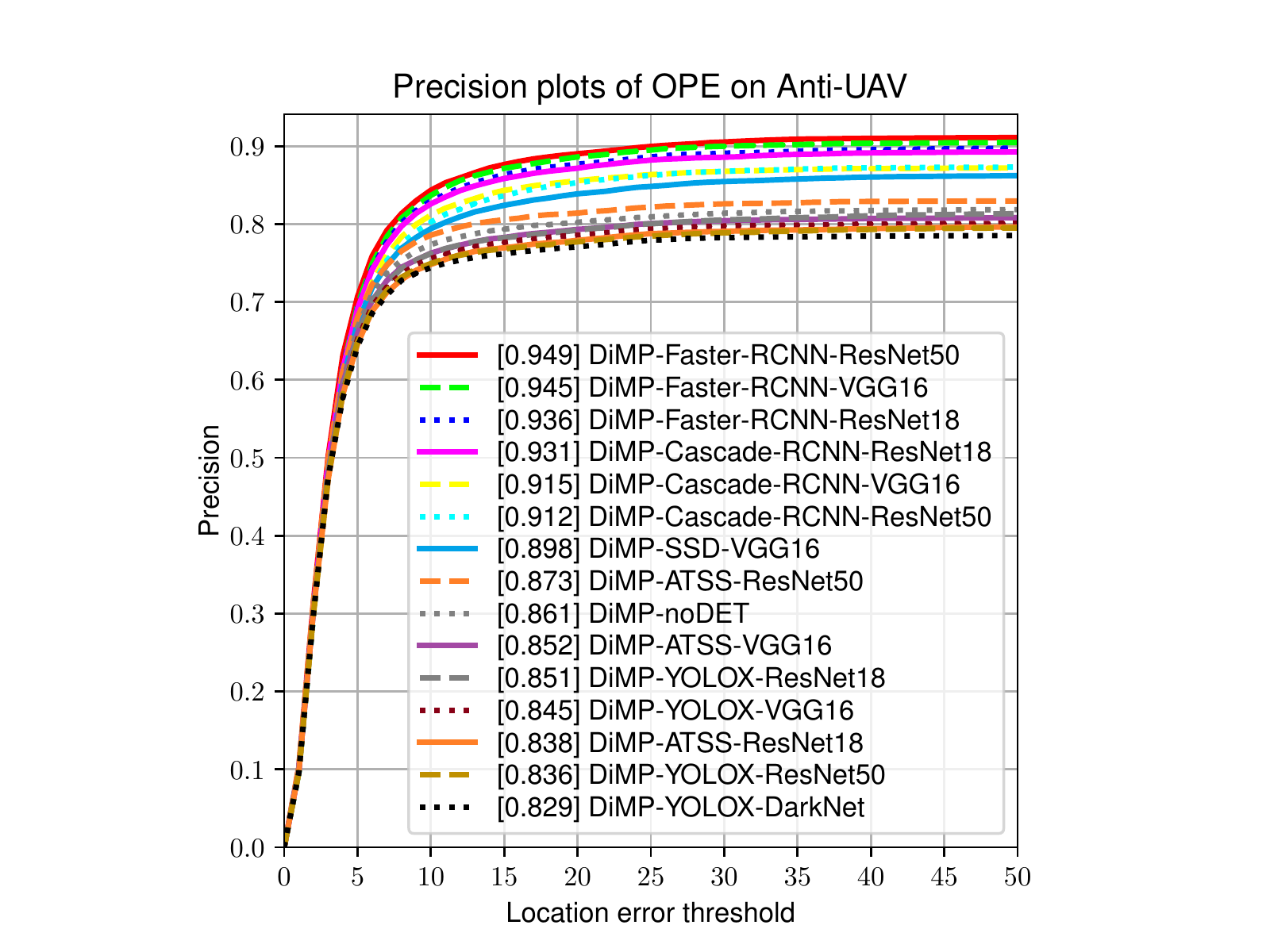}} 
    \subfigure[LTMU]{\includegraphics[width=0.24\textwidth]{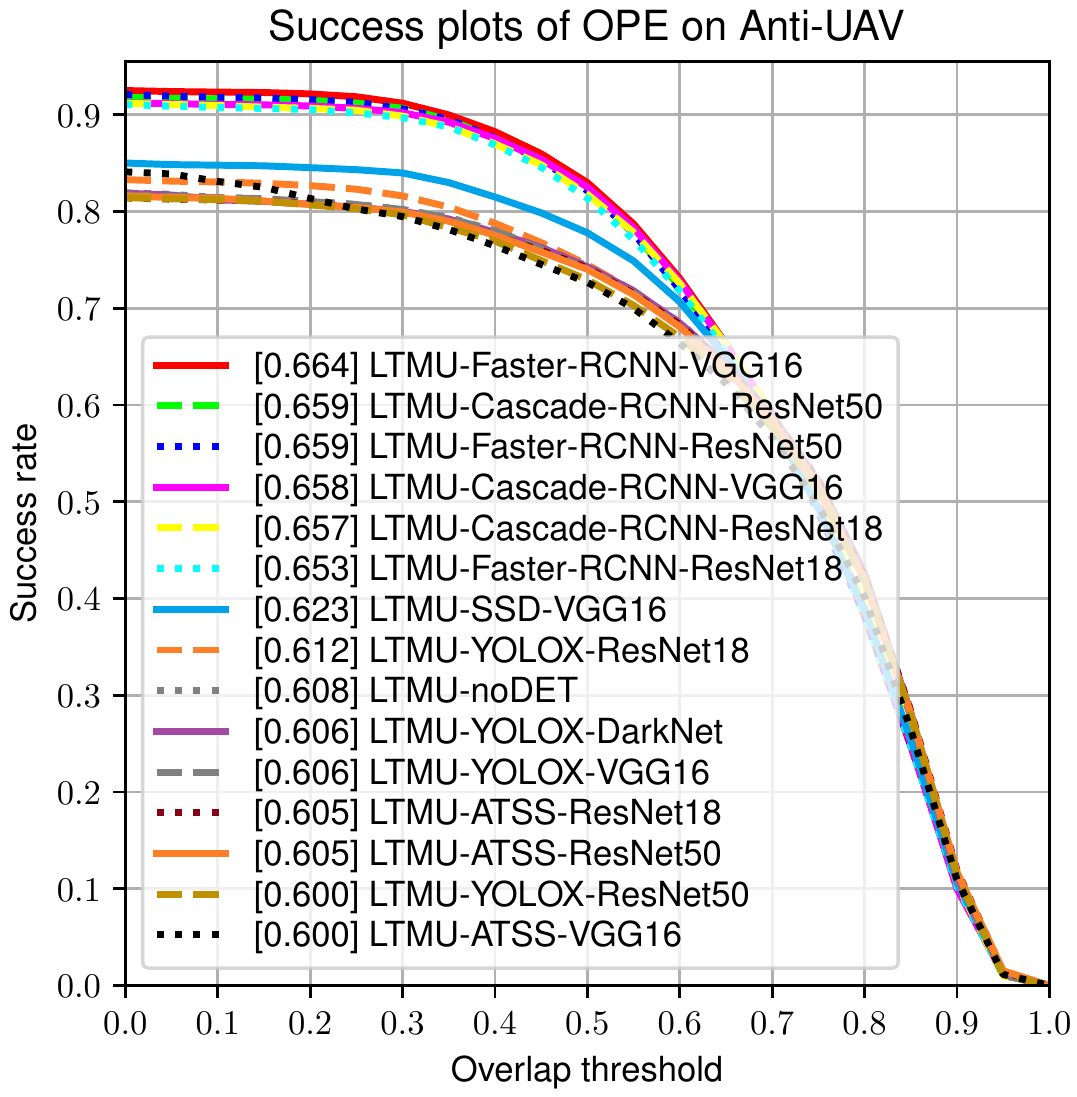}
    \includegraphics[width=0.24\textwidth]{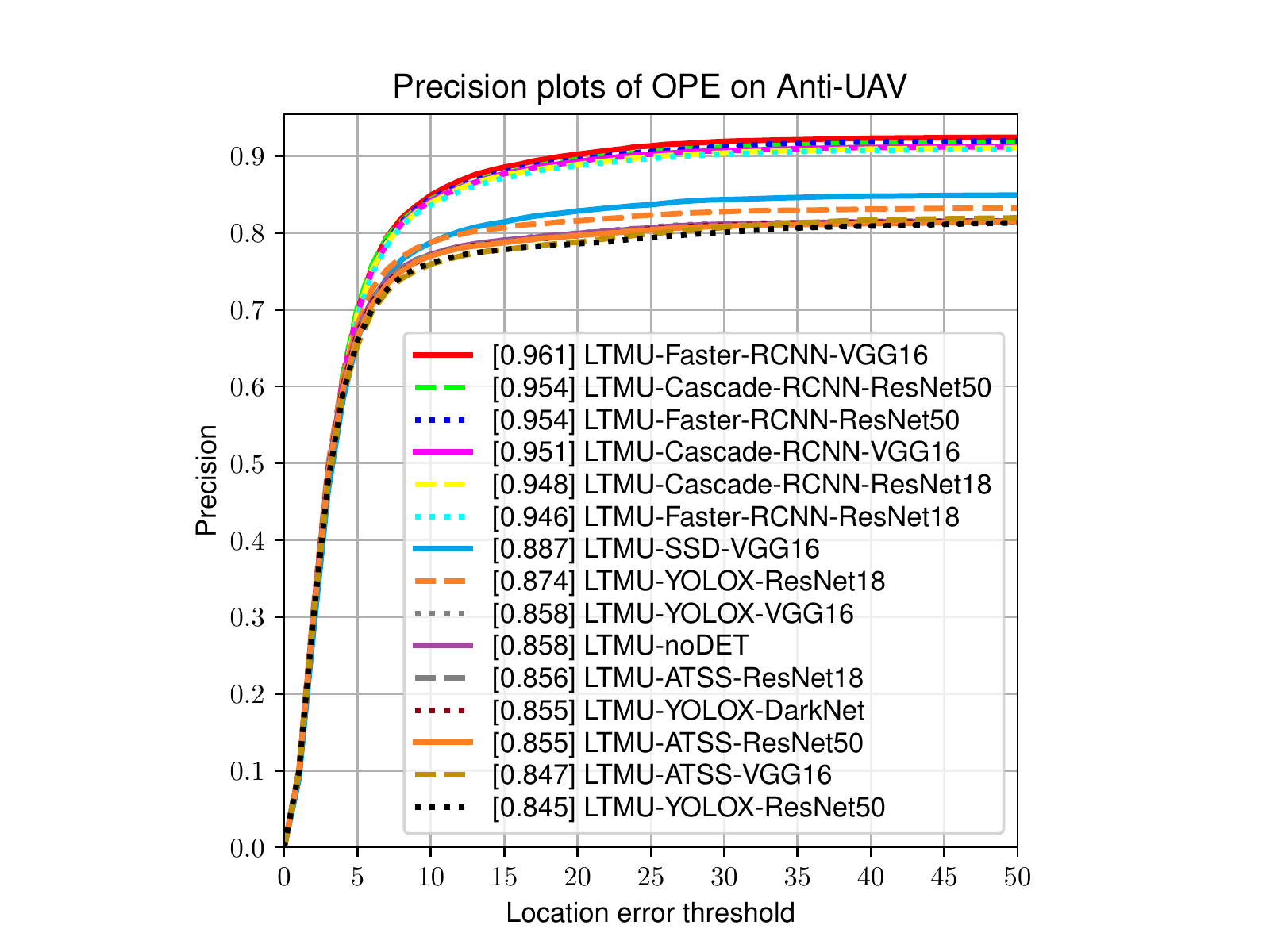}} 
    \caption{Success and precision plots of trackers on the DUT Anti-UAV dataset.}
    \label{fig:success-plot}
\end{figure*}

\begin{figure*}
    \centering
    \includegraphics[width=0.98\textwidth]{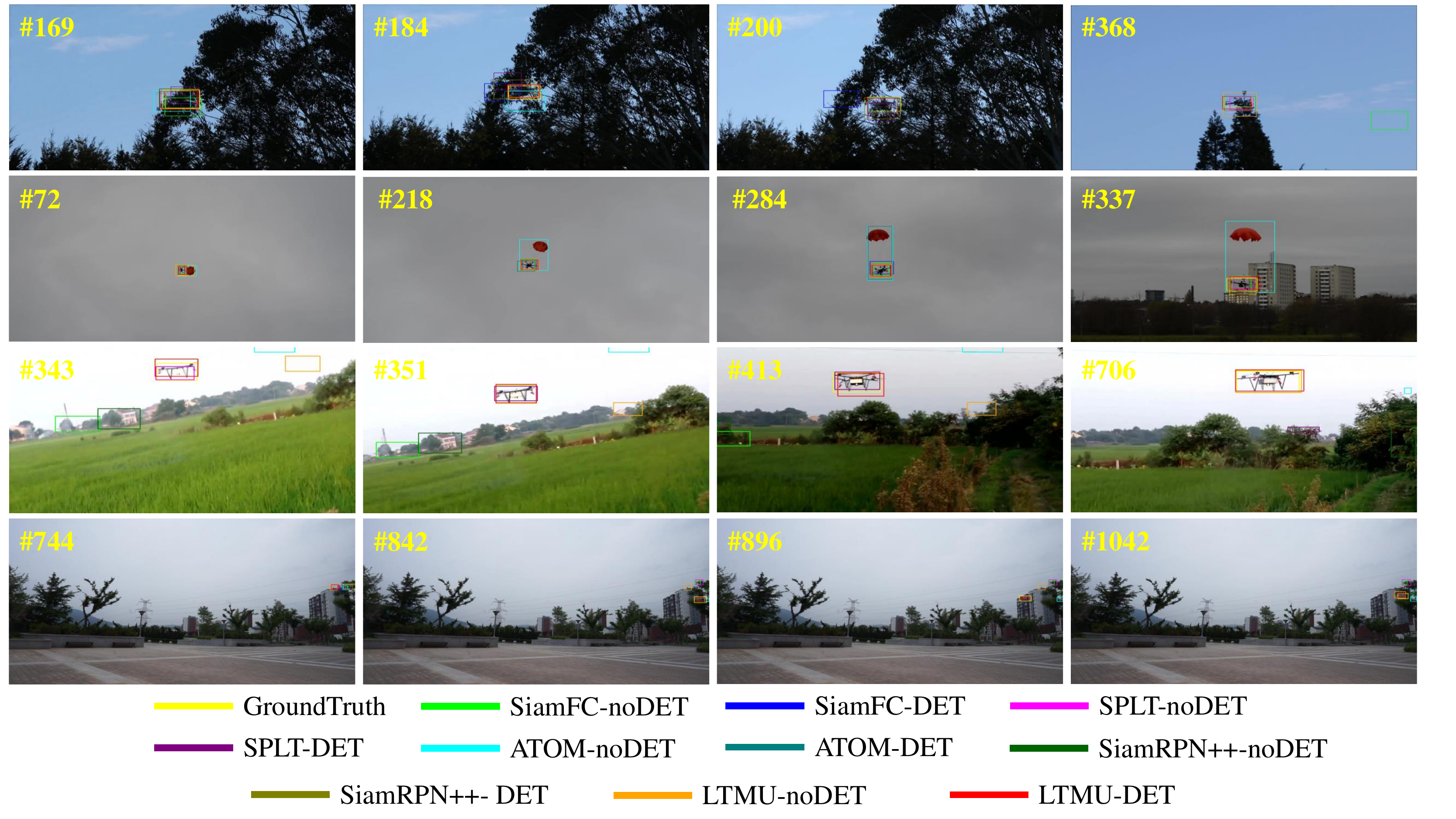}
    \caption{Qualitative comparison of tracking results. "noDET" means pure tracking results without fusing detection, and we choose the model Faster-RCNN-VGG16 for "DET" results. Better viewed in color with zoom-in.}
    \label{fig:tracking_bbox_plot}
\end{figure*}

On the basis of the proposed fusion strategy, we try different combinations of a series of trackers and detectors. Specifically, we select the eight aforementioned trackers (SiamFC, ECO, SPLT, ATOM, SiamRPN++, DiMP, TransT and LTMU) and five detectors with different types of backbone networks (14 different versions in total). The detailed tracking results are shown in Table~\ref{tab:tracker_detector_results}. The success and precision plots of each tracker are shown as Fig.~\ref{fig:success-plot}. After fusing detection, the tracking performance of all trackers is improved significantly. For instance, compared with the baseline tracker SiamFC, the fused method SiamFC+Faster-RCNN(VGG16) increases by 23.4\% in terms of the Success. The best-performing tracker LTMU also improves its performance further after fusing the detector Faster-RCNN(VGG16). The degree of tracking performance improvement depends on the detection algorithm. For most trackers, Faster-RCNN is a better fusion choice, especially for its VGG16 version. On the contrary, ATSS hardly provides additional performance benefits to trackers. Aside from Faster-RCNN, Cascade-RCNN can also enhance the tracking performance. Fig.~\ref{fig:tracking_bbox_plot} shows the qualitative comparison of the original trackers and our fusion methods, where the model Faster-RCNN-VGG16 is chosen as the fused detector. After fusing detection, trackers can perform better in most scenarios. Among them, the performance of LTMU-DET is best that can handle most challenges.
\begin{algorithm}
\caption{Fusion strategy of trackers and detectors}
\label{alg:combination}
\begin{algorithmic}[1]
\REQUIRE ~~ $\mathcal{T}$, $\mathcal{D}$, $I_{\{0,N-1\}}$, $GT_0$
\ENSURE ~~ $R_{\{0,N-1\}}$
\STATE $\mathcal{T}$.init($I_0$,$GT_0$);\\
\STATE $R_0 = GT_0$;\\
\FOR{each $t \in [1,N-1]$}
    \STATE $bbox_t$, $score_t$ = $\mathcal{T}$.track($I_t$);
    \IF{$score_t<\tau_t$}
        \STATE $bboxes_d$, $scores_d$ = $\mathcal{D}$.detect($I_t$);
        \IF{$length(bboxes_d)>0$}
            \STATE $score_d$ = Max($scores_d$);\\ 
            \STATE Collect the corresponding $bbox_d$ from $bboxes_d$;\\
            \IF{$score_d>\tau_d$ and $score_d>score_t$}
                \STATE $R_t=bbox_d$;
            \ELSE
                \STATE $R_t=bbox_t$;
            \ENDIF
        \ENDIF
    \ENDIF
\ENDFOR
\end{algorithmic}
\end{algorithm}


\section{Conclusion}
In this paper, we propose the DUT Anti-UAV dataset for UAV detection and tracking. It contains two sets, namely, detection and tracking. The former has 10,109 objects from 10,000 images, which are split into three subsets (training, testing, and validation). The latter contains 20 sequences whose average length is 1240. All images and frames are annotated manually and precisely. We set 14 different versions of the detectors from the combination of 5 types of detection algorithms and 3 types of backbone networks. These methods are retrained using our detection-training dataset and evaluated on our detection-testing dataset. Besides, we present the tracking results of the 8 trackers on our tracking dataset. To improve the tracking performance further and make full use of our detection and tracking datasets, we propose a simple and clear fusion strategy with trackers and detectors and evaluate the tracking results of the combinations of 8 trackers and 14 detectors. Extensive experiments show that our fusion strategy can improve the tracking performance of all trackers significantly.\par

\bibliographystyle{IEEEtran}
\bibliography{egbib}

\end{document}